\documentclass[letterpaper, 10 pt, conference]{ieeeconf}  % Comment this line out if you need a4paper

\IEEEoverridecommandlockouts                              % This command is only needed if
\usepackage{cite}
\usepackage{amsmath,amssymb,amsfonts}
\usepackage{graphicx}
\usepackage{textcomp}
\usepackage{color}
\usepackage{xcolor}
\usepackage{xcolor,pifont}
\usepackage{booktabs}
\usepackage{lipsum}
\usepackage[linesnumbered,ruled,vlined]{algorithm2e}
\usepackage{algpseudocode} 
\usepackage{subcaption}

\usepackage{caption}
\DeclareCaptionFormat{twolinetab}{#1#2\\#3}
\SetKwInput{KwData}{Global params.}
\SetKw{return}{return}
\SetKw{Break}{break}

\usepackage{amsmath,amsfonts,amssymb,bm}

\usepackage{siunitx}
\DeclareSIUnit \parsec {pc}
\DeclareSIUnit \electronvolt {eV}
\DeclareSIUnit \pixel {px}
\DeclareSIUnit \arcmin {arcmin}
\DeclareSIUnit \erg {erg}
\DeclareSIUnit \joul {J}
\DeclareSIUnit \Bq {Bq}
\DeclareSIUnit \byte {B}

\usepackage{tikz}
\usepackage{pgfplots}
\pgfplotsset{compat=1.14}
\usetikzlibrary{backgrounds,arrows,automata,shapes,positioning,calc,through,spy}
\pgfdeclarelayer{background}
\pgfdeclarelayer{foreground}
\pgfsetlayers{background,main,foreground}

\usepackage{url}

\usepackage{breakurl}
\usepackage[breaklinks]{hyperref}

\usepackage[nolist,nohyperlinks]{acronym}
\acrodef{gps}[GPS]{Global Positioning System}
\acrodef{slam}[SLAM]{Simultaneous Localization And Mapping}
\acrodef{slams}[SLAMs]{Simultaneous Localization And Mapping systems}
\acrodef{gps}[GPS]{Global Positioning System}
\acrodef{rtk}[RTK]{Real-time Kinematics}
\acrodef{gnss}[GNSS]{Global Navigation Satellite System}
\acrodef{ros}[ROS]{Robot Operating System}
\acrodef{api}[API]{Application Programming Interface}
\acrodef{uav}[UAV]{Uncrewed Aerial Vehicle}
\acrodef{ugv}[UGV]{Uncrewed Ground Vehicle}
\acrodef{uv}[UV]{Ultra-Violet}
\acrodef{led}[LED]{Light-emitting Diode}
\acrodef{mbzirc}[MBZIRC]{Mohamed Bin Zayed International Robotics Challenge}
\acrodef{darpa}[DARPA]{Defense Advanced Research Projects Agency}
\acrodef{imu}[IMU]{Inertial Measurement Unit}
\acrodef{lti}[LTI]{Linear time-invariant}
\acrodef{mpc}[MPC]{Model Predictive Control}
\acrodef{uvdar}[UVDAR]{Ultra-Violet Direction And Ranging}
\acrodef{dof}[DOF]{degree-of-freedom}
\acrodef{dofs}[DOFs]{degrees-of-freedom}
\acrodef{lidar}[LiDAR]{Light Detection and Ranging}
\acrodef{esc}[ESC]{Electronic Speed Controller}
\acrodef{lkf}[LKF]{Linear Kalman Filter}
\acrodef{ukf}[UKF]{Unscented Kalman Filter}
\acrodef{ekf}[EKF]{Extended Kalman Filter}
\acrodef{ras}[RAS]{Robotics and Automation Society}
\acrodef{ieee}[IEEE]{Institute of Electrical and Electronics Engineers}
\acrodef{mrs}[MRS]{Multi-robot Systems Group}
\acrodef{ue}[UE]{Unreal Engine}
\acrodef{ue5}[UE5]{Unreal Engine 5}
\acrodef{hla}[HLA]{High-Level Autonomy}
\acrodef{hitl}[HITL]{Hardware-In-The-Loop}
\acrodef{pc}[PC]{Point Cloud}

\newcommand{\refsec}[1]{Sec.~\ref{#1}}

\newcommand{\minus}{\scalebox{0.75}[1.0]{$-$}}

\def\YES{\ding{52}}
\def\NO{\ding{55}}

\newcommand{\PREPRINTYEAR}{2025}
\newcommand{\PUBLISHEDIN}{IEEE International Conference on Robotics and Automation}
\usepackage[placement=top,vshift=-7]{background}
\SetBgScale{0.8}
\SetBgContents{\copyright{} \PREPRINTYEAR~\PUBLISHEDIN. PREPRINT VERSION - DO NOT DISTRIBUTE.}
\SetBgColor{black}
\SetBgAngle{0}
\SetBgOpacity{1.0}

\usepackage{tikz}

\newcommand\copyrighttext{%
  \footnotesize \textcopyright 2025 IEEE.  Personal use of this material is permitted.  Permission from IEEE must be obtained for all other uses, in any current or future media, including reprinting/republishing this material for advertising or promotional purposes, creating new collective works, for resale or redistribution to servers or lists, or reuse of any copyrighted component of this work in other works.}
\newcommand\copyrightnotice{%
\begin{tikzpicture}[remember picture,overlay]
\node[anchor=south,yshift=10pt] at (current page.south) {\fbox{\parbox{\dimexpr\textwidth-\fboxsep-\fboxrule\relax}{\copyrighttext}}};
\end{tikzpicture}%
}

\title{\LARGE \bf FlightForge: Advancing UAV Research with Procedural Generation of High-Fidelity Simulation and Integrated Autonomy\\
}

\author{
  David \v{C}apek,
  Jan Hrn\v{c}\'{i}\v{r},
  Tom\'{a}\v{s} B\'{a}\v{c}a,
  Jakub Jirkal,
  Vojt\v{e}ch Von\'{a}sek,
  Robert P\v{e}ni\v{c}ka and
  Martin Saska
  \thanks{The authors are with the Multi-robot Systems Group, Faculty of Electrical
  Engineering, Czech Technical University in Prague, Czech Republic (\protect\url{http://mrs.felk.cvut.cz/}).
  This work has been supported by the Czech Science Foundation (GAČR) under research project No. 23-06162M, by the European Union under the project Robotics and Advanced Industrial Production (reg. no. CZ.02.01.01/00/22\_008/0004590) and by CTU grant no SGS23/177/OHK3/3T/13. %% <-this % stops a space
  Computational resources were provided by the e-INFRA CZ project (ID:90254). %supported by the Ministry of Education, Youth and Sports of the Czech Republic.
  }}

\begin{document}

% \thispagestyle{empty}
% \onecolumn
% {
%   \topskip0pt
%   \vspace*{\fill}
%   \centering
%   \LARGE{%
%     \copyright{} \PREPRINTYEAR~\PUBLISHEDIN\\\vspace{1cm}
%     Personal use of this material is permitted.
%     Permission from \PUBLISHEDIN~must be obtained for all other uses, in any current or future media, including reprinting or republishing this material for advertising or promotional purposes, creating new collective works, for resale or redistribution to servers or lists, or reuse of any copyrighted component of this work in other works.}
%     \vspace*{\fill}
% }
% \NoBgThispage
% \twocolumn          	
% \BgThispage

  \maketitle
  \copyrightnotice
  \thispagestyle{empty}
  \pagestyle{empty}

  \begin{abstract}
    Robotic simulators play a crucial role in the development and testing of autonomous systems, particularly in the realm of Uncrewed Aerial Vehicles (UAV).
    However, existing simulators often lack high-level autonomy, hindering their immediate applicability to complex tasks such as autonomous navigation in unknown environments.
    This limitation stems from the challenge of integrating realistic physics, photorealistic rendering, and diverse sensor modalities into a single simulation environment.
    At the same time, the existing photorealistic UAV simulators use mostly hand-crafted environments with limited environment sizes, which prevents the testing of long-range missions.
    This restricts the usage of existing simulators to only low-level tasks such as control and collision avoidance.
    To this end, we propose the novel FlightForge UAV open-source simulator.
    FlightForge offers advanced rendering capabilities, diverse control modalities, and, foremost, procedural generation of environments.
    Moreover, the simulator is already integrated with a fully autonomous UAV system capable of long-range flights in cluttered unknown environments.
    The key innovation lies in novel procedural environment generation and seamless integration of high-level autonomy into the simulation environment.
    Experimental results demonstrate superior sensor rendering capability compared to existing simulators, and also the ability of autonomous navigation in almost infinite environments. 
  \end{abstract}

  %%{ Introduction

  \section*{Software and additional multimedia materials}

  \noindent
  \textbf{Code:} \href{https://github.com/ctu-mrs/flight_forge}{https://github.com/ctu-mrs/flight\_forge}\\
  \textbf{Multimedia:} \href{https://mrs.felk.cvut.cz/flight-forge}{https://mrs.felk.cvut.cz/flight-forge}

  \section{Introduction}

  Simulators are commonly used in robotics for development, initial testing, or parameter tuning~\cite{erez2015simulation}.
  These tasks would be too expensive (both from a monetary and time perspective) or impractical with real robots.
  Failures in simulations are not fatal in comparison to the failures of real robots~\cite{liu2021role}.
  Simulators can also decrease the gap for researchers to pioneer their ideas for, e.g., autonomous flight with \acp{uav}.
  For example, the recent progress in learning-based methods in robotics, such as champion-level racing drone~\cite{kaufmann2023champion}, or deployment of learned policies for quadrupedal robots~\cite{miki2022learning}, has been mostly driven by high-fidelity physics simulations.
  Yet, existing UAV simulators do not incorporate a fully autonomous \ac{uav} system and procedural generation of environments, which are key factors to simulate long-range missions with autonomous sensor-based navigation.

  %%{ INTRO FOTO

  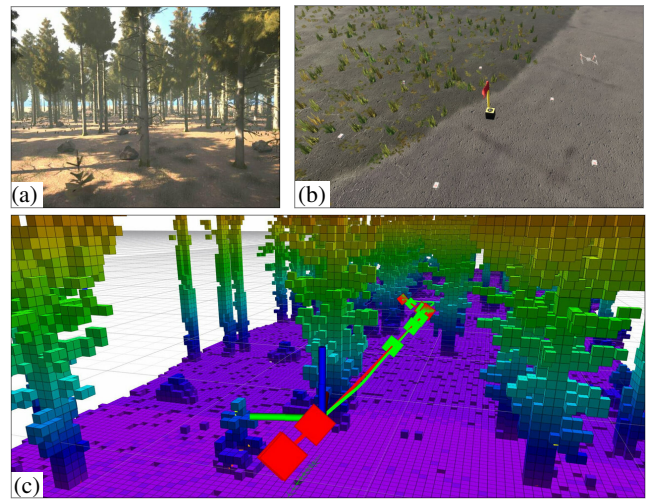
\begin{figure}[t]
    \centering
    \hspace{-5pt}\subfloat {\begin{tikzpicture}
      \node[anchor=south west,inner sep=0] (a) at (0,0) {\includegraphics[width=0.20\textwidth]{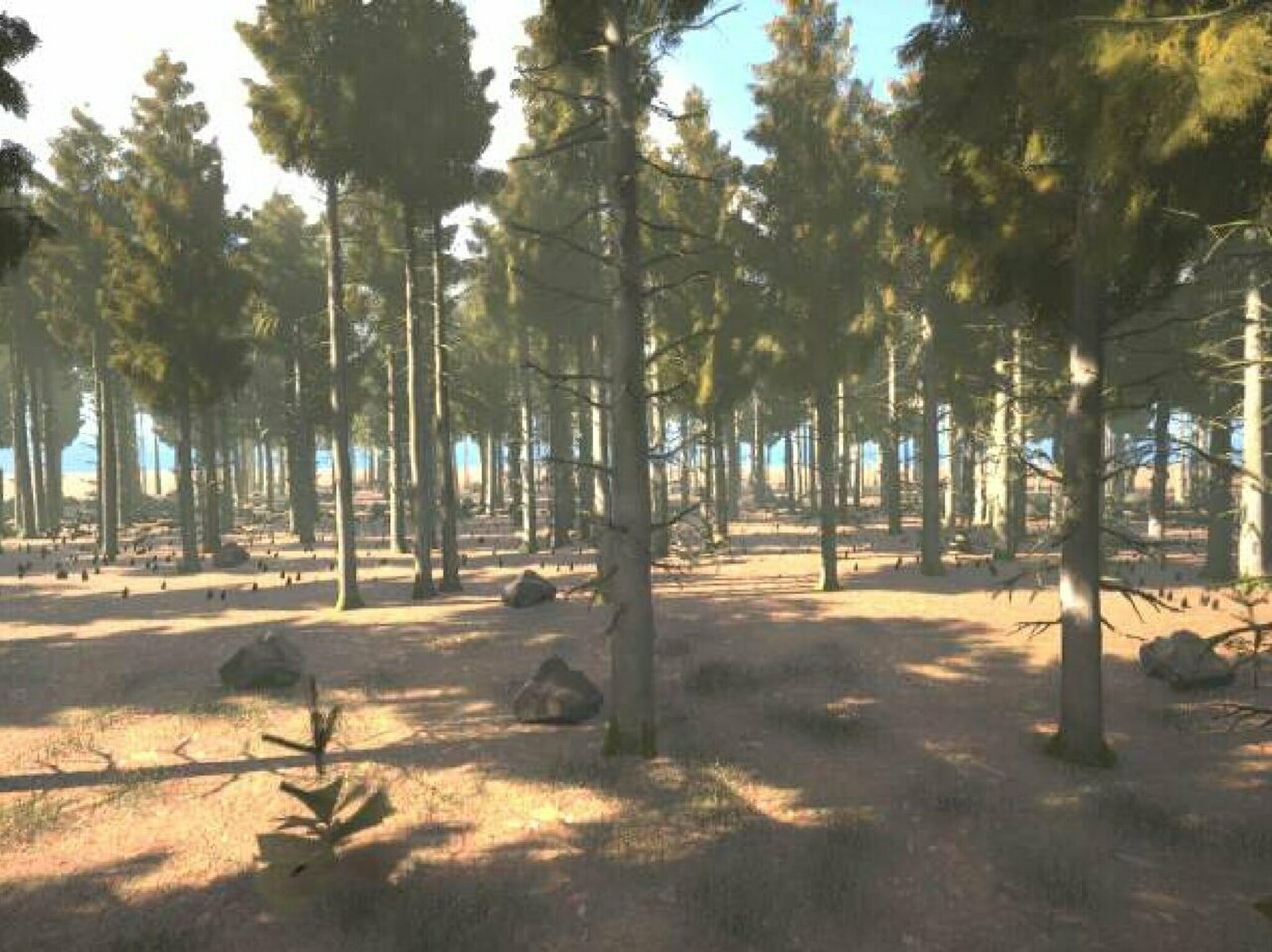}};
      \begin{scope}[x={(a.south east)},y={(a.north west)}]
        \fill[white] (0.001, 0.001) rectangle (0.12,0.13);
        \fill[draw=black, draw opacity=0.5, fill opacity=0] (0,0) rectangle (1, 1);
        \draw (0.05,0.06) node [text=black] {\small (a)};
      \end{scope}
    \end{tikzpicture}
    }%
    \hspace{0pt}\subfloat {\begin{tikzpicture}
      \node[anchor=south west,inner sep=0] (a) at (0,0) {\includegraphics[width=0.26\textwidth]{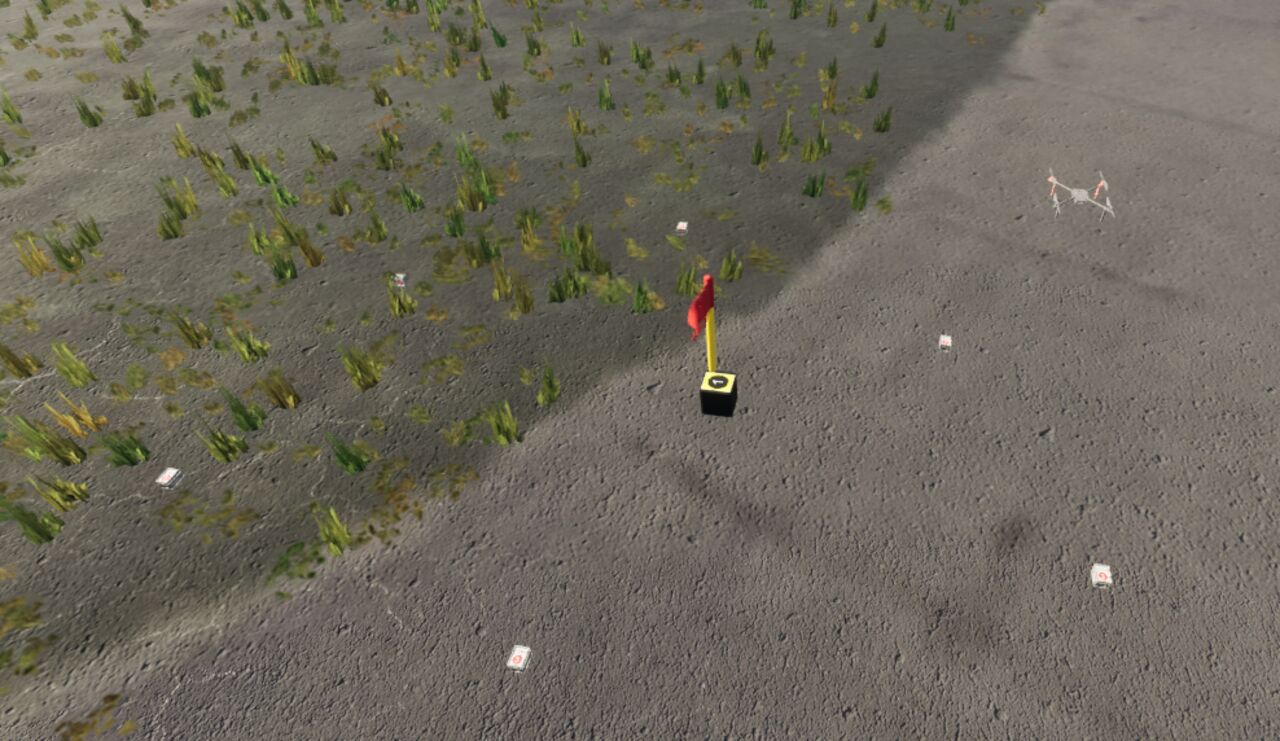}};
      \begin{scope}[x={(a.south east)},y={(a.north west)}]
        \fill[white] (0.001, 0.001) rectangle (0.1,0.13);
        \fill[draw=black, draw opacity=0.5, fill opacity=0] (0,0) rectangle (1, 1);
        \draw (0.05,0.06) node [text=black] {\small (b)};
      \end{scope}
    \end{tikzpicture}
    }\\
    \subfloat {\begin{tikzpicture}
      \node[anchor=south west,inner sep=0] (a) at (0,0) {\includegraphics[width=0.473\textwidth,trim={0 1cm 0 2cm}, clip]{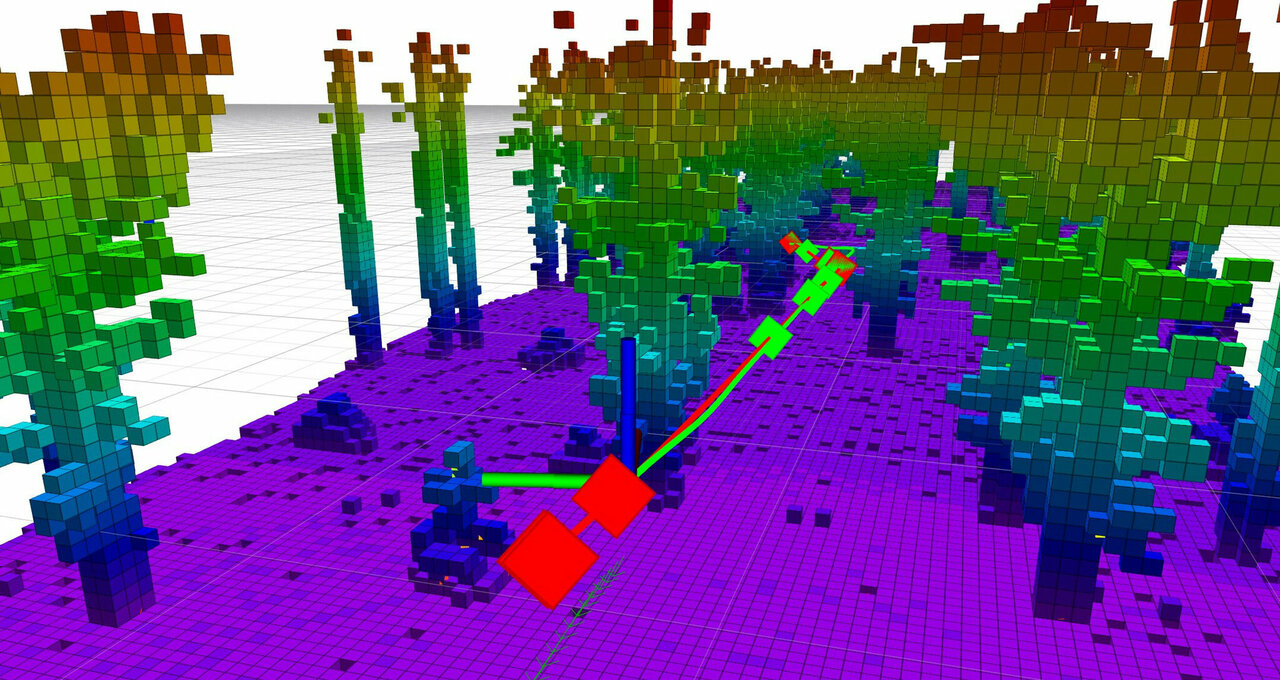}};
      \begin{scope}[x={(a.south east)},y={(a.north west)}]
        \fill[white] (0.001, 0.001) rectangle (0.05,0.09);
        \fill[draw=black, draw opacity=0.5, fill opacity=0] (0,0) rectangle (1, 1);
        \draw (0.025,0.04) node [text=black] {\small (c)};
      \end{scope}
    \end{tikzpicture}
    \vspace{-0.8em}
    }
    \caption{Example of FlightForge functionalities: a) image from the UAV's onboard camera, b) the third-person view of the UAV during a simulation of the Sprin-D Fully Autonomous Flight Challenge in 2024, c) visualization of the 3D obstacle mapping and planning pipeline provided by the integrated MRS UAV System \cite{baca2021mrs}. Demonstration videos are available at \href{https://mrs.felk.cvut.cz/flight-forge}{mrs.felk.cvut.cz/flight-forge}.}
    \label{fig:intro_fig}
    \vspace{-1.5em}
  \end{figure}

  %%}

  With the growing applications of UAV systems in various fields, there is also a growing need for high-fidelity simulations.
  Photorealistic sensor rendering is especially important as UAVs are typically equipped with a camera, and considering various light sources, textures, shades, and reflections is important for realistic sensor simulation.
  Besides, tools for editing photorealistic environments are necessary.
  Several high-fidelity robotic simulators have already been based on game engines~\cite{madaan2020airsim, song2021flightmare}, e.g., Unity and Unreal Engine, as these engines offer advanced rendering features.
  Yet, these were mostly proposed for learning flight~\cite{song2021flightmare} with Reinforcement Learning, or to support deep learning data collection~\cite{madaan2020airsim}.

  The available high-fidelity simulators lack \ac{hla} navigation features, such as \ac{uav} control, estimation, trajectory generation, path planning, mapping and localization.
  Due to the missing \ac{hla}, users of simulators often focus on developing these features, instead of solving the tasks that are the true focus of their research, e.g., multi-UAV cooperation or multi-modal mapping.
  Existing simulators also do not support procedural environment generation and thus the users have to prepare the environments (terrain, building, obstacles, etc.) manually, which is not suitable for making large-scale environments.
  However, in the Sprin-D Fully Autonomous Flight Challenge in 2024, which motivated this work, the teams were tasked to fly without GPS over a course of several kilometers autonomously.
   We propose novel FlightForge simulator based on \ac{ue5} which is directly connected to the fully autonomous MRS UAV System\footnote{\href{https://github.com/ctu-mrs/mrs_uav_system}{github.com/ctu-mrs/mrs\_uav\_system}}~\cite{baca2021mrs,hert2023mrs}.
  Users can utilize publicly available \ac{ue5} assets (e.g., trees, terrains, furniture) when creating environments, thereby saving time on detailed modeling.
  Moreover, to the best of our knowledge this is the fist UAV simulator that proposes to use procedural generation of large-scale terrains, which is based on several parameters (e.g., terrain roughness or forest density).
  Compared to the existing works \cite{song2021flightmare,madaan2020airsim}, the barebone simulator offers various levels of control modality of the UAVs ranging from high-level position commands to the low-level single rotor throttle commands.
 On the other hand, the simulator, when connected to the MRS UAV System, allows fully autonomous flight in unknown environment.
  The proposed open-sourced simulator\footnote{\href{https://github.com/ctu-mrs/flight_forge}{github.com/ctu-mrs/flight\_forge}} supports single and multi-\ac{uav} deployment.
  The \acp{uav} can be equipped dynamically with RGB and RGB-D cameras and a 3D \ac{lidar}, all of them incoming at realistic data rates.
  All sensor modalities are accompanied by annotated semantic segmentation ground truth.
  The intensity channel of the \ac{lidar} sensor is also supported.
  FlightForge also supports the \ac{hitl} mode, where the real \ac{uav} is receiving the sensor data from the simulator.
  This was utilized by the authors in the Sprin-D Challenge for testing the vision-based flight in vast simulated environments generated both using procedural generation and existing map data of otherwise restricted military air base the challenge takes place in.
  FlightForge and the related packages from the MRS UAV System are easy-to-install \emph{debian} packages in the Ubuntu OS, or usable as Docker and Apptainer containers.

We show that the rendering speeds achieved by the simulated RGB camera are comparable to, or better than, of the existing simulators~\cite{song2021flightmare}.
The simulated 3D \ac{lidar} sensor is capable of producing up to 32768 points at a frequency of \SI{69}{Hz}.
Finally, we show the integration with a fully autonomous UAV system through the capability of autonomous exploration and mapping of a procedurally generated environment and demonstrate a multi UAV scenario.

  %%}
  
  %%{ Related work
  
  \section{Related Work}

  Robotic simulators are often used in research~\cite{collinsAReview2021}, e.g., for development of methods, and design on new robotic systems~\cite{kriegmanScalablePipelineDesigning2020} or for training learning-based methods~\cite{peng2017deeploco,miki2022learning}.
  General-purpose simulators allow users to simulate wide range of robots (e.g., CoppeliaSim~\cite{coppeliaSim}, Webots~\cite{Webots} or Ignition Gazebo~\cite{koenigDesignUseParadigms2004,gazeboNew}), but they may lack particular features (e.g., support for hardware-in-the-loop the specific robots or high-fidelity rendering).
  In this section, we focus solely on simulators suitable for \acp{uav}.
  We refer to the comprehensive survey~\cite{collinsAReview2021} about physical simulators.
  Simulators of \acp{uav} should support system dynamics, various environments, and basic sensors, including cameras, possibly with more advanced features like weather simulation (e.g., wind, rain, fog)~\cite{dimmig2024survey,Silano2023SurveySimulator}.
  An overview of related simulators is shown in Tab.~\ref{tab:comparison}.

  Ignition Gazebo~\cite{koenigDesignUseParadigms2004,gazeboNew} is a general-purpose physics-based simulator.
  Gazebo uses OpenGL for rendering but lacks photorealistic rendering.
  It offers cameras and \acp{lidar} and can be extended by plugins to simulate GPS and barometer~\cite{meyerComprehensiveSimulationQuadrotor2012}.
  Gazebo features a set of predefined robots (e.g., DJI Mavic 2 PRO).
  UAVs can be controlled using PX4 and Ardupilot SITL flight-controllers.
  Several other simulators are based on Gazebo,
  e.g., the Hector Quadrotor package~\cite{meyerComprehensiveSimulationQuadrotor2012}, RotorS~\cite{furrerRotorSModularGazebo2016}
  and CrazyS~\cite{silano2018crazys}.
  RotorS~\cite{furrerRotorSModularGazebo2016} is a rudimentary \ac{uav} simulator  
  providing several multirotor UAV models (e.g., AscTec Hummingbird \& Firefly) and  
  provides \ac{imu} and 3D pose for the \ac{uav}.
  Besides, it provides a simulation of the VI-Sensor. 
  This package also contains some example controllers, basic worlds, and a joystick interface. 

  Isaac Sim~\cite{isaacsim} is a photorealistic high-fidelity simulator for various robotic platforms (mobile, legged, manipulators).
  The above mentioned simulators~\cite{koenigDesignUseParadigms2004,gazeboNew,meyerComprehensiveSimulationQuadrotor2012,furrerRotorSModularGazebo2016,silano2018crazys,isaacsim} either lack the high-fidelity rendering, or do not provide high-level autonomy capabilities.

\begin{table}
\centering
{\small
\setlength{\tabcolsep}{1pt}
\caption{\label{tab:comparison}
Key features of UAV simulators.
Sensors are: I: IMU, G: GPS, SS: semantic segmentation, L:\ac{lidar},
R+D: both RGB image + depth image, 
$\diamond$: \ac{lidar} + intensity \ac{lidar},
$\dagger:$ not open-source in the time of writing}
\vspace{-8pt}
  \begin{tabular}{l@{\hspace{0pt}}cccc@{\hspace{-1pt}}c}
\toprule
Simulator & Rendering & Physics & Sensors & Autonomy \\
 \midrule
\cite{Webots} Webots
 & WREN  
 & ODE% \cite{russel2008ODE} 
% & IMU, GPS, RGB, \ac{lidar}
& I,G,RGB,L
   & \NO              \\
\cite{isaacsim} Isaac Sim     
 & Omniverse                
 & PhysX %~\cite{physx}                
% & IMU, GPS, RGB, Depth, SS. 
& I,G,R+D,SS
  & \NO              \\
\cite{guerra2019flighgoggles} FlightGoggles
 & Unity %~\cite{juliani2020unity}               
 & Flexible             
 %& IMU, RGB                                             
& I,RGB
   & \NO              \\
\cite{madaan2020airsim} AirSim
 & UE 4          
 & PhysX %~\cite{physx}   
 %& IMU, RGB+D,  SS.
 & I,R+D,SS
 & \NO              \\
\cite{song2021flightmare} Flightmare
 & Unity %~\cite{juliani2020unity}                    
 & Flexible             
% & IMU, RGB+D,  SS.                         
& I,R+D,SS
  & \NO              \\
\cite{cui2024fastsim}  Fastsim & Unity & Flexible & I,R+D,SS,L & \YES $\dagger$ \\
 \midrule
 \textbf{FlightForge} & 
 \textbf{UE 5} & \textbf{Flexible}    
 & \textbf{I,G,R+D,SS,L$\diamond$} & \YES   \\
 \bottomrule
\end{tabular}
      
}
\vspace{-2em}
\end{table}

  The open-source AirSim~\cite{madaan2020airsim} supports research in AI, computer vision, and learning-based approaches for autonomous vehicles (cars and UAVs), for which it offers many sensors  
  and supports software-in-the-loop and hardware-in-the-loop simulation
  with flight controllers (e.g., PX4 and ArduPilot).
  AirSim is based on the \ac{ue} 4 providing photorealistic rendering.
  AirSim can also emulate weather effects (e.g., rain, snow, fog).

  Flightmare is an open-source flexible modular quadrotor simulator~\cite{song2021flightmare} based on the Unity game engine.
  Its central principle is the decoupling of a rendering and physics engine, 
  so the users can decide which physics engine will be running.
  The simulator provides an RGB-D camera with ground-truth depth and semantic segmentation, rangefinder and supports collision detection between the UAVs and the environment.
  Cameras can be further modified, users can change the field of view, focal length, or even lens distortion.
  Flightmare can simulate up to several hundred of agents in parallel, which is useful, e.g., in learning-based research.

  FlightGoggles~\cite{guerra2019flighgoggles} focuses on photorealistic
  rendering; it even generates simulated environments 
  using photogrammetry (i.e., by reconstructing 3D shapes (meshes) from many photos) to enable realistic simulation of exteroceptive sensors like RGB-D cameras.
  The simulator is based on Unity, and it adopts the modular architecture where other parts (e.g., sensor simulation or collision detection) are implemented as modules.
  The 3D meshes are generated with different levels of detail for fast rendering and collision detection.
  FlightGoggles nodes and the API can be used with either ROS~\cite{quigley2009ros} or LCM~\cite{Moore2009LightweightCA}.
  FlightGoggles can have moving obstacles or light, which can be controlled in real-time.

  FastSim~\cite{cui2024fastsim} is a recent simulator for UAVs based on Unity.
  It supports IMU, RGB+D, segmentation, and event cameras.
  Moreover, it claims to provide high-level autonomy modules (control, motion planning and mapping modules).
  However, in the time of writing this manuscript, FastSim is still not published as an open-source.

  %%}

  %%{ Methodology

  \section{Methodology}

\begin{figure}
    \centering
    \includegraphics[width=1\columnwidth, trim={0 0 0 0}, clip]{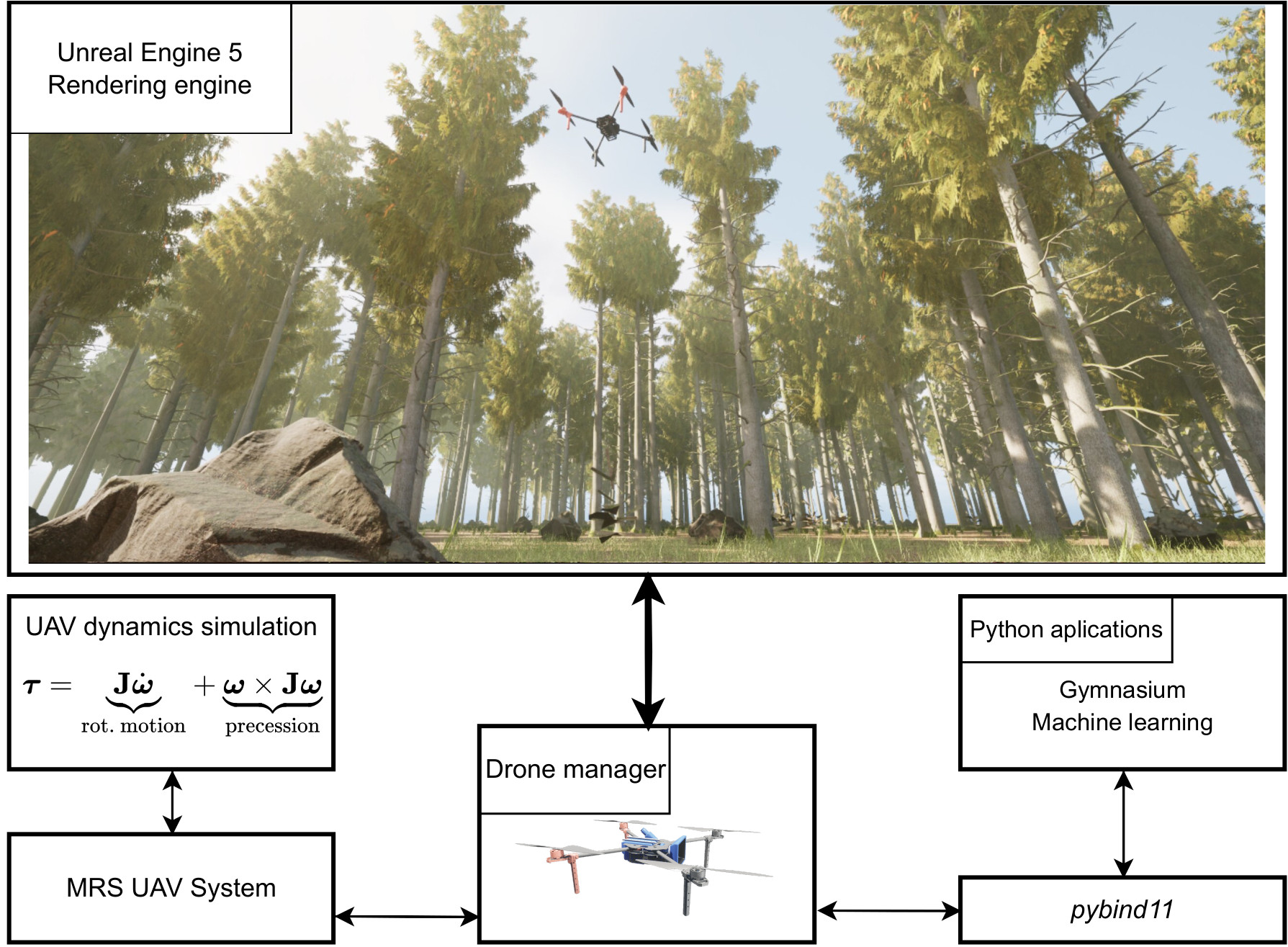}
    \vspace{-1.8em}
    \caption{The schematic of the architecture of the simulator.\label{fig:architecture}}
    \vspace{-2em}
  \end{figure}

  FlightForge consists of three primary components: the realistic rendering engine, the \ac{uav} dynamics simulator and the connector for the MRS \ac{uav} System.
  The components are loosely coupled, which allows the \ac{uav} dynamics simulator to achieve higher simulation rates and to be used independently.
  The interface for the MRS \ac{uav} System is provided, which allows the user to connect the simulator to the ROS ecosystem and to use the ROS tools for the development of the control algorithms and sensor processing.
  Furthermore, an interface is provided for Python applications like Gymnasium, allowing the user to use the simulator to develop reinforcement learning solutions with the simulator.
  The architecture of the simulator overview is shown in Fig.\,\ref{fig:architecture}.

  %%{ rendering engine
  
  \subsection{Rendering engine}
  
  FlightForge is based on the state-of-the-art \ac{ue5} game engine. 
  \ac{ue5} is distinguished by delivering high-quality rendering of the virtual environment, including realistic lighting, shadows, and textures.
  The simulator offers two types of environments: indoor and outdoor, which can be used for various tasks such as navigation, mapping, and localization.
  New environments can be created in the \ac{ue} editor, which allows the user to create the environment from scratch or use publicly available assets from the \ac{ue} marketplace.
  For the creation of the \emph{Forest} environment, the procedural generation component of \ac{ue5} was used to generate the environment.

  \subsection{Procedural terrain generation}
  Due to the rising need to develop, test, and deploy UAVs in long-term and long-range missions, simulation should provide an efficient way to support
  flight in large-scale environments.
  While nowadays simulators use manually predefined scenarios, we 
  utilize procedural generation to dynamically create the environment.

\begin{figure}
    \centering
    \includegraphics[width=0.8\columnwidth]{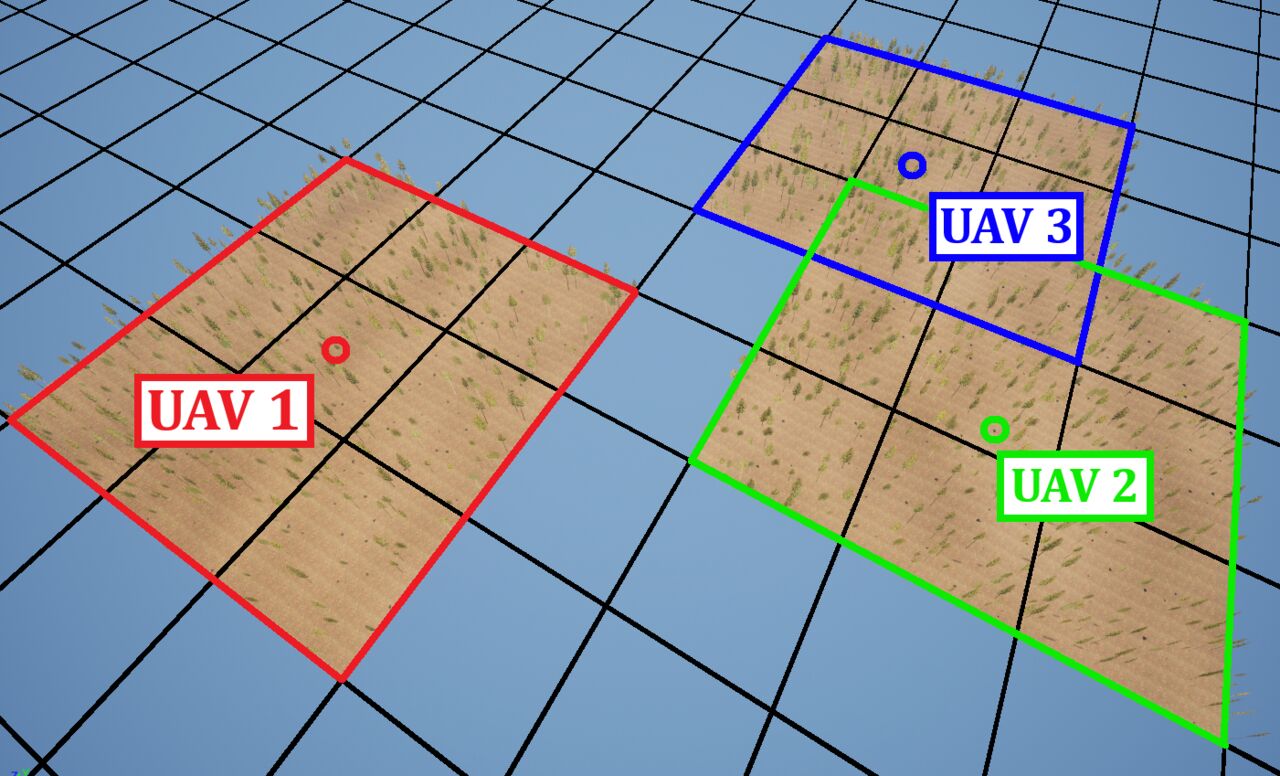}
    \vspace{-0.5em}
    \caption{Cells that are not within a visibility range from the spectator and drones (blue) are removed to save computational resources. \label{fig:infiniteforest}
    }
    \vspace{-1.5em}
  \end{figure}

  The environment is divided into rectangular cells (each contains
  meshes for the terrain, trees, obstacles, etc.); each cell has 8-direct neighbors.
  The cells are dynamically created (and removed) based on their visibility
  from the spectator's and UAVs' positions, i.e., only visible ones are rendered (see Fig.~\ref{fig:infiniteforest}) to reduce computational and memory resources.
  The terrain type is controlled by parameters such as terrain roughness and forest density, 
  and it is generated using Perlin noise~\cite{lagae2010survey,perlin1985animage}.
  The terrain is represented as a 3D triangular mesh, and the position
  of each vertex is generated by the Perlin noise based on the position
  of the vertex and the terrain parameters.
  Foliage, including trees and grass, is generated using \ac{ue5} PCG tool, which samples the terrain surface and filters points to achieve the desired density.
  The maintenance of terrain cells is shown in Alg.~\ref{alg:terrainGeneration}; this 
  algorithm is called in the main loop of the simulator in each simulation step.

\vspace{-1.0em}
\begin{algorithm}
\caption{Procedural terrain generation\label{alg:terrainGeneration}}
\KwIn{$P$: positions of all UAVs, $s$: position of the spectator}
\KwData{$C$ existing terrain cells}
\KwOut{updated terrain cells $C$}
\hrule    
{\small
$C_{new} = \emptyset$\;
\For{$p \in P \cup \{s\}$}{
    $cell\_idx$ = getCellBasedOnPosition(p)\;
    $neighbors\_idx$ = getNeighborCells($cell\_idx$)\;
    \For{$cell \in neighbors\_idx \cup \{cell\_idx\} $}{
        \If{isVisible($cell, p$)}{
            \If{$cell \notin C$}{
                createNewTerrainAtCell($cell$)\;
                createFoliagePositionsAtCell($cell$)\;
                filterFoliagePointsBasedOnDensity($cell$)\;
                placeTreesAndGrassAtCell($cell$)\;
            }
            $C_{new} = C_{new} + \{ cell \}$\;
        }
    }
    \For{$cell \in C$}{
        \If{not isVisible($cell,p$)}{
            removeCellAndStopRenderingIt($cell$)\;
        }
    }
}
$C = C_{new}$\tcp*{new terrain cells}
%\return $C$\;
}
\end{algorithm}
\vspace{-1.5em}

  %%}

  %%{ uav dynamics modelling
  
  \subsection{UAV Dynamics modeling}
  \label{sec:uav_dynamics_modeling}
  
  The rendering engine is loosely coupled with a standalone underactuated multirotor UAV dynamics simulation.
  The dynamics simulator is responsible for simulating the translational and rotational motion of the UAV body, and the force generation process of a virtual propulsion system.
  Each propeller is approximated using the simplified \emph{qudratic thrust model} 
  \begin{equation}
    F \equiv k \omega^2 \label{eq:propeller_thrust_model},\ \dot{\omega} = \minus\left(\omega - \omega_d\right)/\tau_m,
  \end{equation}
  where $F$ is the thrust force, $k$ is a thrust coefficient, and $\omega$ is the propeller's angular velocity.
  The propeller velocity dynamics is modeled as a first-order system with $\dot{\omega}$ being the angular acceleration, $\tau_m$ the time constant\footnote{The default motor time constant is a user parameter and was chosen as \SI{30}{\milli\second} by default.}, $\omega$ the current angular velocity and $\omega_d$ the desired angular velocity.
  The desired angular velocity $\omega_d$ of each propeller is the most low-level user-definable input.
  
  The forces produced by all motors result in intrinsic torques and a collective thrust action on the UAV body in the body frame.
  For a quadrotor X configuration, the input-output relationship between these quantities is defined by the \emph{force-torque} allocation matrix:
  \begin{equation}
    \begin{bmatrix}
      F_t\\
      \tau_1\\
      \tau_2\\
      \tau_3
    \end{bmatrix} = \underbrace{\begin{bmatrix}
      1 & 1 & 1 & 1 \\
      -d/\sqrt{2} & d/\sqrt{2} & d/\sqrt{2} & -d/\sqrt{2} \\
      -d/\sqrt{2} & d/\sqrt{2} & -d/\sqrt{2} & d/\sqrt{2} \\
      -c_{tf} & -c_{tf} & c_{tf} & c_{tf}\\
    \end{bmatrix}}_{\text{$\bm{\Gamma}$, force-torque allocation matrix}} \begin{bmatrix}
      F_1\\
      F_2\\
      F_3\\
      F_4\\
    \end{bmatrix}\label{eq:allocation},
  \end{equation} where $d$ is the diagonal of the frame's geometry and $c_{tf}$ is the torque-generating propeller constant.
  The number of propellers and the \emph{force-torque allocation matrix} is user-definable parameter of the simulation.
  
  The rotational dynamics of the UAV body is first composed out of the Euler's equation of motion
  \begin{equation}
    \bm{\tau} = \underbrace{\mathbf{J}\dot{\bm{\omega}}}_{\text{rot. motion}} + \underbrace{\bm{\omega}\times\mathbf{J}\bm{\omega}}_{\text{precession}},
  \end{equation}
  which describes how the torque $\bm{\tau} = \begin{bmatrix}\tau_1,\tau_2, \tau_3\end{bmatrix}^\intercal$ causes the angular acceleration of the body $\dot{\bm{\omega}}$ factored by the moment of inertia $\mathbf{J}$, and the precession motion.
    Secondly, the intrinsic angular velocity $\bm{\omega}$ is part of the orientation dynamics $\dot{\mathbf{R}} = \mathbf{R}\mathbf{\Omega}$, where $\dot{\mathbf{R}}$ is derivative of the rotation matrix, $\mathbf{R} \in SO(3)$ is the 3D orientation, and $\mathbf{\Omega} \in \mathbf{R}^{3 \times 3}$ such that $\mathbf{\Omega}\,\mathbf{v}= \bm{\omega}\times\mathbf{v}, \forall \mathbf{v} \in \mathbb{R}^3$ is the tensor of angular velocity.
  
\begin{figure}
    \centering
      \includegraphics[width=0.3\textwidth]{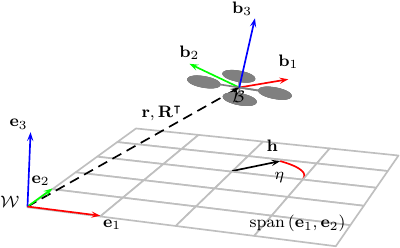}
      \vspace{-1em}
      \caption{
        Illustration of world-fixed frame $\mathcal{W}$ = $\{\mathbf{e}_1$, $\mathbf{e}_2$, $\mathbf{e}_3\}$ with the 3D position and the orientation of the \ac{uav} body.
        The body-fixed frame $\mathcal{B}$ = $\{\mathbf{b}_1$, $\mathbf{b}_2$, $\mathbf{b}_3\}$ relates to $\mathcal{W}$ by the translation $\mathbf{r} = \left[x, y, z\right]^{\intercal}$ and by rotation $\mathbf{R}^{\intercal}$.
        The \ac{uav} heading vector $\mathbf{h}$, which is a projection of $\hat{\mathbf{b}}_1$ to the plane $span\left(\mathbf{e}_1, \mathbf{e}_2\right)$, forms the heading angle $\eta = \mathrm{atan2}\left(\mathbf{b}_1^\intercal\mathbf{e}_2, \mathbf{b}_1^\intercal\mathbf{e}_1\right) = \mathrm{atan2}\left(\mathbf{h}_{(2)}, \mathbf{h}_{(1)}\right)$.
        }
        \label{fig:coordinate_frames}
        \vspace{-1em}
    \end{figure}

    The translational dynamics from the $2^{\text{nd}}$ Newton's law is:
    \begin{equation}
      \ddot{\mathbf{r}}^{\mathcal{W}} = \frac{1}{m} \mathbf{R} {\begin{bmatrix}0, 0, F_t\end{bmatrix}^\intercal}^{\mathcal{B}} + \mathbf{g}^{\mathcal{W}},
    \end{equation}
    where $\ddot{\mathbf{r}}^{\mathcal{W}}$ is the acceleration of the UAV's body in the \emph{world-fixed frame}, $m$ is the UAV's mass, the collective thrust force $F_t$ produced by the propellers acting along the $\bm{b_3}$ axis of the body-fixed frame, and $\mathbf{g}^{\mathcal{W}}$ is the gravity-caused acceleration expressed in the \emph{world-fixed frame}.
    Figure\,\ref{fig:coordinate_frames} illustrates the mentioned coordinate frames.
  
  %%}

    %%{ uav dynamics simulation
    
    \subsection{UAV dynamics simulation}
    \label{sec:uav_dynamics_simulation}
    
    The UAV dynamics presented in \refsec{sec:uav_dynamics_modeling} are discretized into a set of discrete nonlinear differential equations and are forward integrated in real time using a Runge-Kutta4.
    The system is implemented in C\texttt{++} and it leverages the Boost Odeint library, which provides very fast implementation of the Runge-Kutta4.
    All the dynamics parameters are user-configurable, including the step size, which is set to $1/250 s$ by default.
    The state of the dynamics system is regularly passed to the Unreal Engine renderer, which reflects the UAV's $\mathbf{r}$ and $\mathbf{R}$ in the engine's world.
    
    The dynamics simulator is also usable in a standalone mode without the rendering counterpart.
    This feature is useful for the design and testing of automatic control methods.
    The standalone simulator is released as a separate package\footnote{\href{https://github.com/ctu-mrs/mrs_multirotor_simulator}{github.com/ctu-mrs/mrs\_multirotor\_simulator}} as it is part of the core of the MRS UAV System.
    The simulator allows real-time simulation of hundreds of \acp{uav}, which is especially useful for swarm research.
    The dynamics simulation of a single \ac{uav} is provided in the form of a header-only library, which can be copy-pasted into third-party code.
    The single-UAV dynamics simulation can be executed at a rate of above \SI{10}{\kilo\hertz}, which makes it especially useful for learning-based control research.
    
    The standalone simulation allows the user to use any of the following control modalities: (a) individual actuators' throttle, (b) normalized control groups (pitching, rolling, yawing, collective throttle), (c)  body-frame attitude rate $\mathbf{\omega}$ and collective throttle $T$, (d) attitude $\mathbf{R}$ and collective throttle $T$, (e) acceleration $\ddot{\mathbf{r}}$ and heading $\eta$, (f) acceleration $\ddot{\mathbf{r}}$ and heading rate $\dot{\eta}$, (g) velocity $\dot{\mathbf{r}}$ and heading $\eta$, (h) velocity $\dot{\mathbf{r}}$ and heading rate $\dot{\eta}$, and (i) position $\mathbf{r}$ and heading $\eta$.
    The desired throttle of a motor is a normalized desired angular rate of the motor, i.e., $\omega_d/\omega_{\text{max}}$, where $\omega_{\text{max}}$ is the maximum angular rate of the motor.
    The same applies to the desired collective throttle, which is an arithmetic mean of all the motors' desired angular velocities.
    The dynamics model for the simulated UAV is fully configurable.
    
    %%}

    %%{ sensors
    
    \subsection{Sensors}
    
    The dynamics simulator directly provides the simulation of a 3D accelerometer and 3D gyroscope.
    Within the abstraction layer between the simulator and the MRS UAV System, additional virtual sensors, such as magnetometer, barometer, and \ac{gnss}, are emulated.
    All the sensors can be easily altered by the user with artificial noise and other signal degradation.

    We have developed and implemented a comprehensive suite of virtual sensors within FlightForge, closely mimicking those commonly used in aerial robotics. 
    These sensors, built upon the capabilities of \ac{ue5}, include a camera, a depth camera, and a \ac{lidar} sensor, with accompanying ground truth semantic segmentation.
    
    \subsubsection{RGB Camera}
    
    The camera sensor creates a high-resolution snapshot of the current  scene, and the user can configure it (e.g., resolution, field of view, and frame rate).
    
    \subsubsection{Depth camera}
    
    The depth camera sensor provides the depth representation of the environment.
    It is fabricated using a stereo camera model, producing a depth map from the disparity of the stereo images.
    This is a superior method for creating a depth camera compared to the grayscale depth cameras that are used in existing simulators, as it provides absolute depth values.
    As with the RGB camera, the user can adjust the resolution, field of view, and frame rate of the depth camera.
    
    \subsubsection{3D \ac{lidar}}
    
    The inclusion of the \ac{lidar} sensor in a drone simulator is of paramount importance.
    The \ac{lidar} sensor provides a 3D \acp{pc} representation of the environment.
    The data is integral for obstacle avoidance, mapping, and localization.
    In the simulator, the \ac{lidar} sensor is emulated using the ray-casting feature.
    The ray casts are preformed in parallel to enhance the performance of the sensor.
    The \ac{lidar} can output intensity \acp{pc}, with configurable intensity values to simulate different sensors. 
    Intensity values are useful for distinguishing different materials, such as grass, trees, and buildings. 
    The adjustable parameters of the \ac{lidar} sensor include the number of rays, the field of view, the sensor's maximum operational range, the update frequency of the sensor, and the noise level.
    
    \subsubsection{Semantic segmentation --- ground truth}
    A ground truth semantic segmentation of the environment is provided by the simulator.
    The semantic segmentation label is assigned to each object in the environment, and the label is used to classify the object.
    The current method allows for a definition of up to 256 different classes of objects in the environments.
    These segmentation labels are fully customizable by the user, allowing for tailored experiments and datasets.
    The label of each object is retrieved either from a segmentation camera or a segmentation \ac{lidar} sensor.
    In the case of the segmentation camera, each pixel is assigned a predefined RGB color, which is then used to classify the object.
    In the case of the segmentation \ac{lidar}, the output is a color \acp{pc} with different RGB colors.
    
    %%}

% HITL %%{
    \section{Hardware-in-the-loop simulation}
    \label{sec:hardware_in_the_loop_simulation}   
    FlightForge, together with the MRS \ac{uav} System, is capable of interfacing with real hardware.
    In the \ac{hitl} mode, FlightForge is connected to a real \ac{uav} which sends the position and orientation data to FlightForge.
    The simulator then uses this data to render the \ac{uav} in the virtual environment and sends the sensory data back to the real \ac{uav}.
    The \ac{hitl} simulation is useful for testing sensory-based navigation and control with real \ac{uav} in a safe environment while getting sensory data from complex or not easily accessible simulated environments.
    This was the case for the Sprin-D Fully Autonomous Flight Challenge, where the competition was held in a restricted area and all prior testing had to be performed in a simulator (see Fig.\,\ref{fig:intro_fig}b and \ref{fig:environments}d).

    % %%}
  
    %%}

    %%{ Results

    \section{Results}

\begin{figure}[htb!]
      \centering
      \subfloat {\begin{tikzpicture}
        \node[anchor=south west,inner sep=0] (a) at (0,0) {\includegraphics[width=0.231\textwidth]{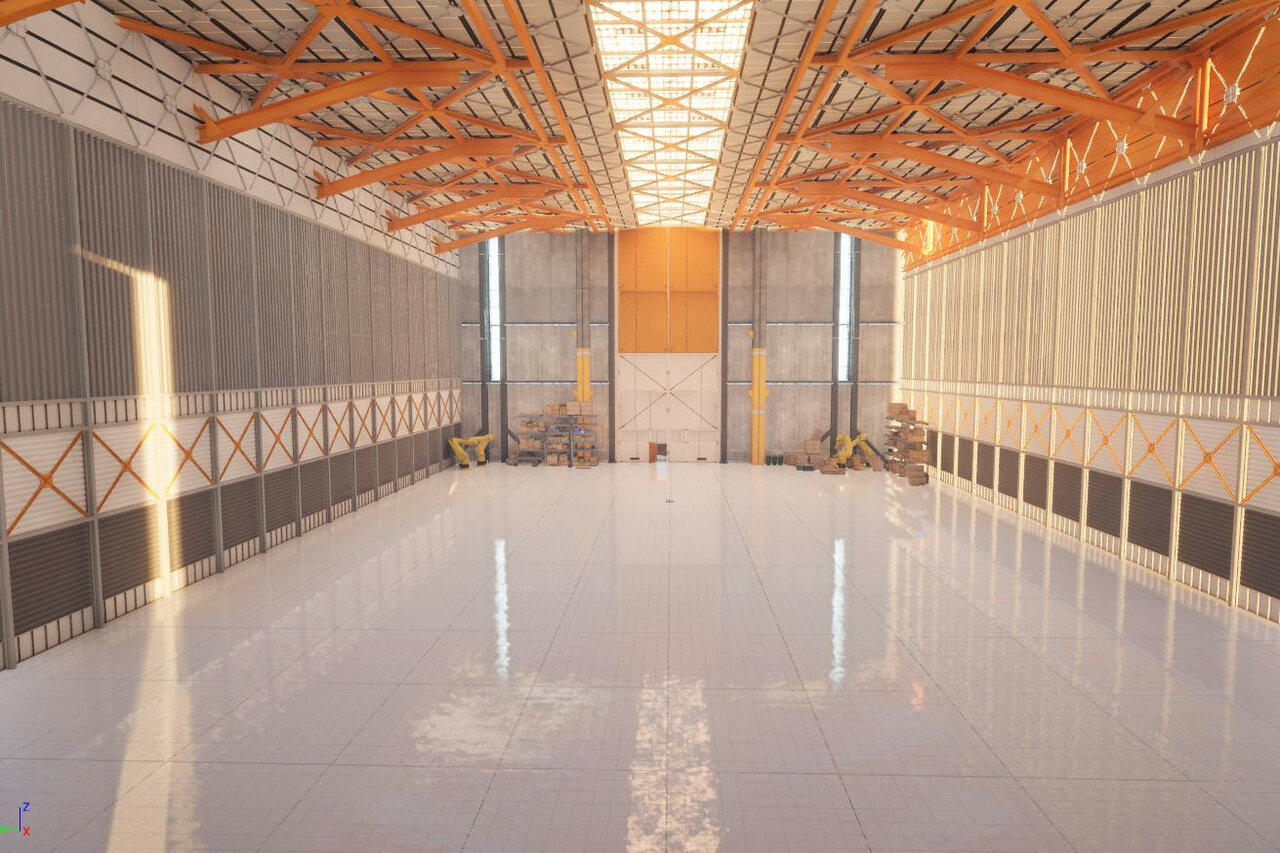}};
        \begin{scope}[x={(a.south east)},y={(a.north west)}]
          \fill[white] (0.001, 0.001) rectangle (0.1,0.13);
          \fill[draw=black, draw opacity=0.5, fill opacity=0] (0,0) rectangle (1, 1);
          \draw (0.05,0.06) node [text=black] {\small (a)};
        \end{scope}
      \end{tikzpicture}
      }%
      \subfloat {\begin{tikzpicture}
        \node[anchor=south west,inner sep=0] (a) at (0,0) {\includegraphics[width=0.231\textwidth]{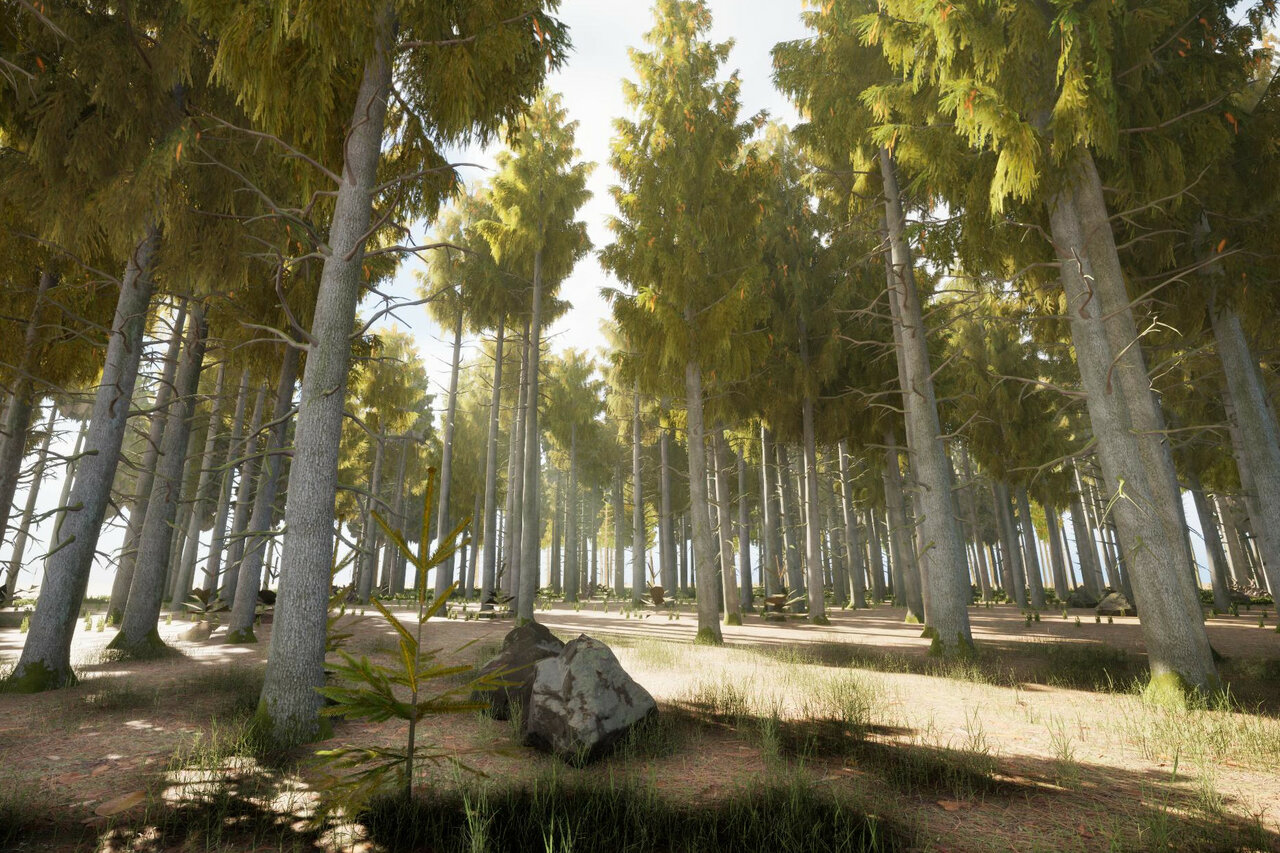}};
        \begin{scope}[x={(a.south east)},y={(a.north west)}]
          \fill[white] (0.001, 0.001) rectangle (0.1,0.13);
          \fill[draw=black, draw opacity=0.5, fill opacity=0] (0,0) rectangle (1, 1);
          \draw (0.05,0.06) node [text=black] {\small (b)};
        \end{scope}
      \end{tikzpicture}
      }\\
      \vspace{-0.1em}
      \subfloat {\begin{tikzpicture}
        \node[anchor=south west,inner sep=0] (a) at (0,0) {\includegraphics[width=0.231\textwidth, trim={0 1cm 0 2cm}, clip]{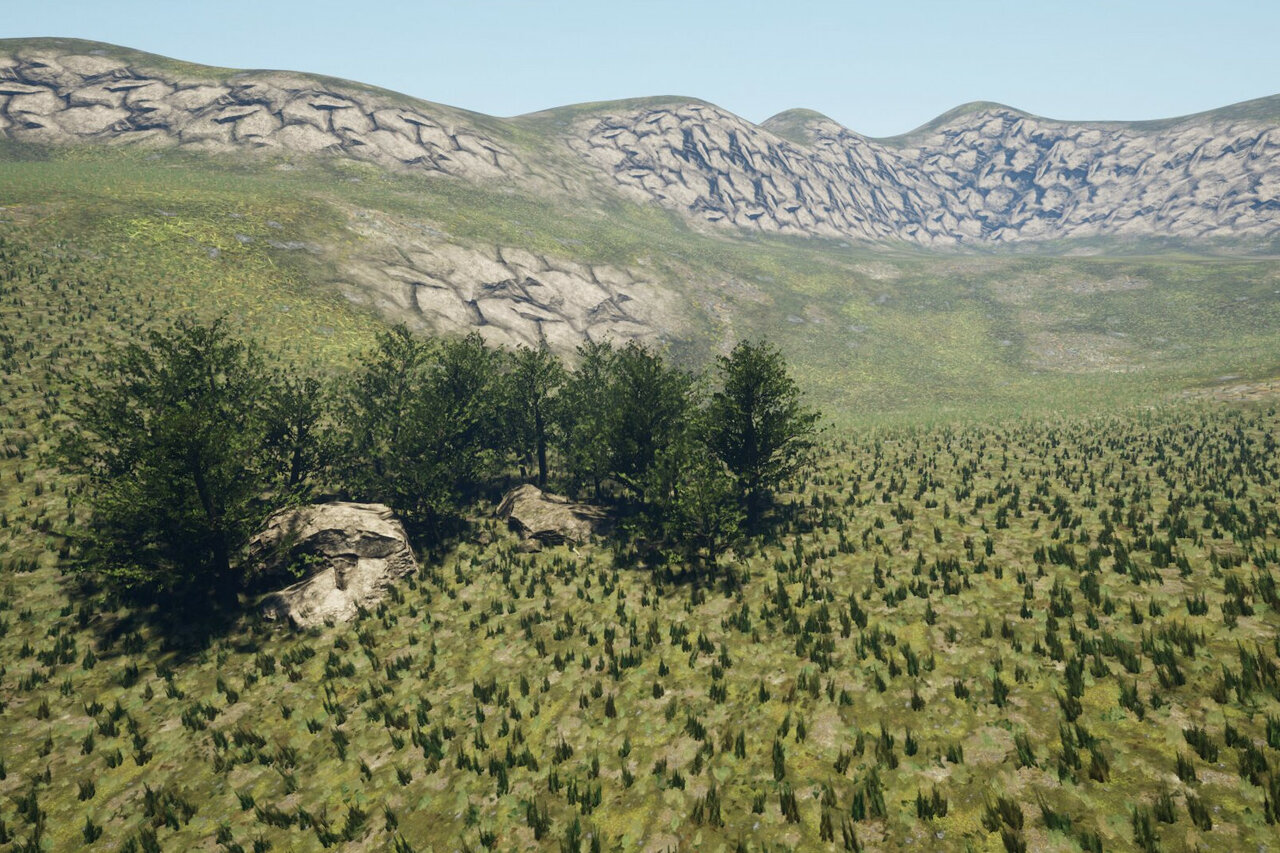}};
        \begin{scope}[x={(a.south east)},y={(a.north west)}]
          \fill[white] (0.001, 0.001) rectangle (0.1,0.22);
          \fill[draw=black, draw opacity=0.5, fill opacity=0] (0,0) rectangle (1, 1);
          \draw (0.05,0.10) node [text=black] {\small (c)};
        \end{scope}
      \end{tikzpicture}
      }%
      \subfloat {\begin{tikzpicture}
        \node[anchor=south west,inner sep=0] (a) at (0,0) {\includegraphics[width=0.231\textwidth, trim={0 1.5cm 0 1.5cm}, clip]{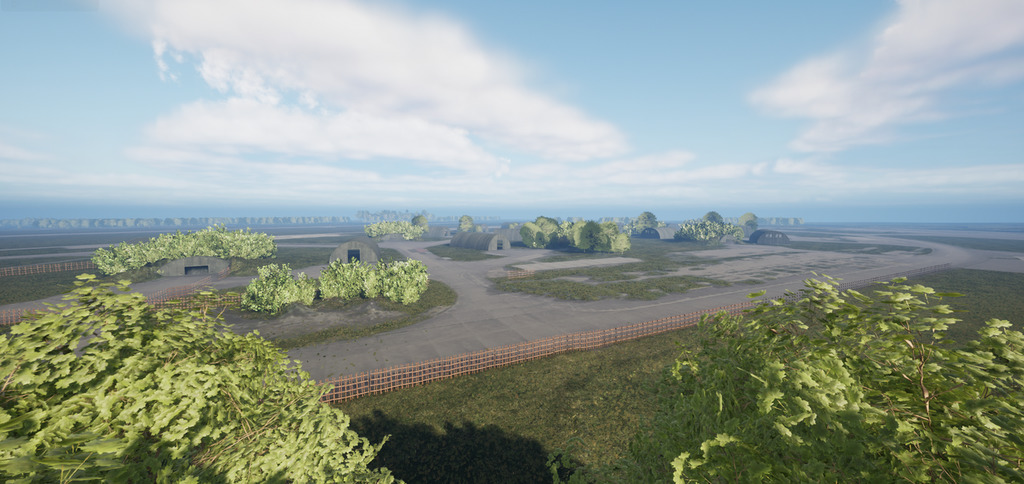}};
        \begin{scope}[x={(a.south east)},y={(a.north west)}]
          \fill[white] (0.001, 0.001) rectangle (0.1,0.22);
          \fill[draw=black, draw opacity=0.5, fill opacity=0] (0,0) rectangle (1, 1);
          \draw (0.05,0.10) node [text=black] {\small (d)};
        \end{scope}
      \end{tikzpicture}
      }% 
      \vspace{-1.1em}
      \caption{Samples of the provided simulation environments: a) \emph{warehouse}, b) \emph{forest}, c) \emph{valley}, d) \emph{Erding air base}.}
      \label{fig:environments}
      \vspace{-0.7em}
    \end{figure}
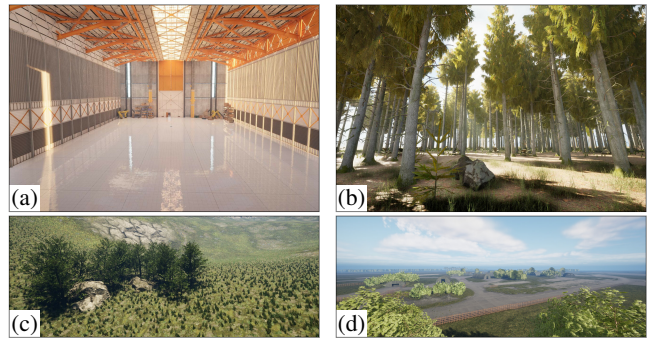

    We show experiments that demonstrate the key capabilities of the FlightForge simulator.
    The experiments are designed to show the performance of the UAV dynamics simulation, the sensor suite, and the integration with the MRS UAV System.
    The experiments were conducted in the \emph{Forest}, \emph{Warehouse}, \emph{Valley}, and \emph{Erding air base} environments, as depicted in Figure\,\ref{fig:environments}.
    Table \ref{tab:comparison} compares the created simulator with other available simulators.
    The experiments were conducted on a laptop with an AMD Ryzen 9 5000HS CPU, 32 GB of RAM, and an Nvidia GeForce RTX 3060 Laptop GPU.

    %%{ TABLE COMPARISON

    %%}

    %%{ FIGURE ENVIRONMENTS

    %%}

    %%{ FIGURE CAMERA OUTPUTS
    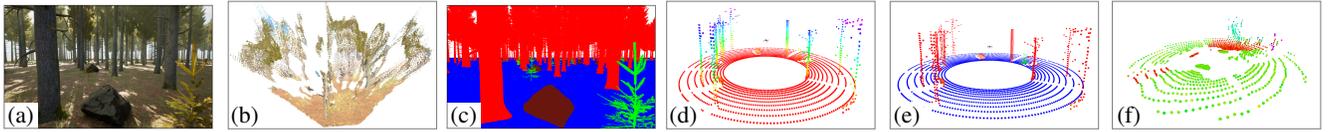
\begin{figure*}[ht!]
      \centering
      \subfloat {\begin{tikzpicture}
        \node[anchor=south west,inner sep=0] (a) at (0,0) {\includegraphics[width=0.155\textwidth]{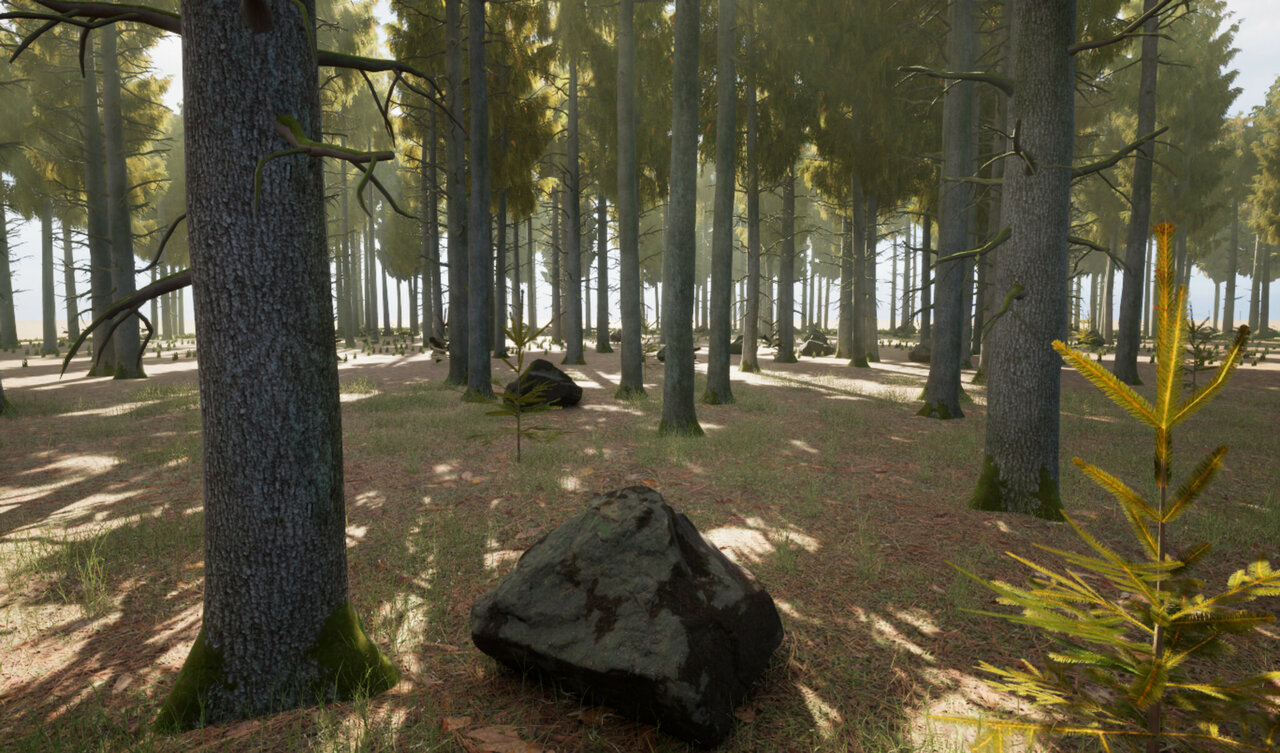} };
        \begin{scope}[x={(a.south east)},y={(a.north west)}]
          \fill[white] (0.001, 0.001) rectangle (0.16,0.21);
          \fill[draw=black, draw opacity=0.5, fill opacity=0] (0,0) rectangle (1, 1);
          \draw (0.08,0.10) node [text=black] {\small (a)};
        \end{scope}
      \end{tikzpicture}
      }%
      \subfloat {\begin{tikzpicture}
        \node[anchor=south west,inner sep=0] (a) at (0,0) {\includegraphics[width=0.155\textwidth]{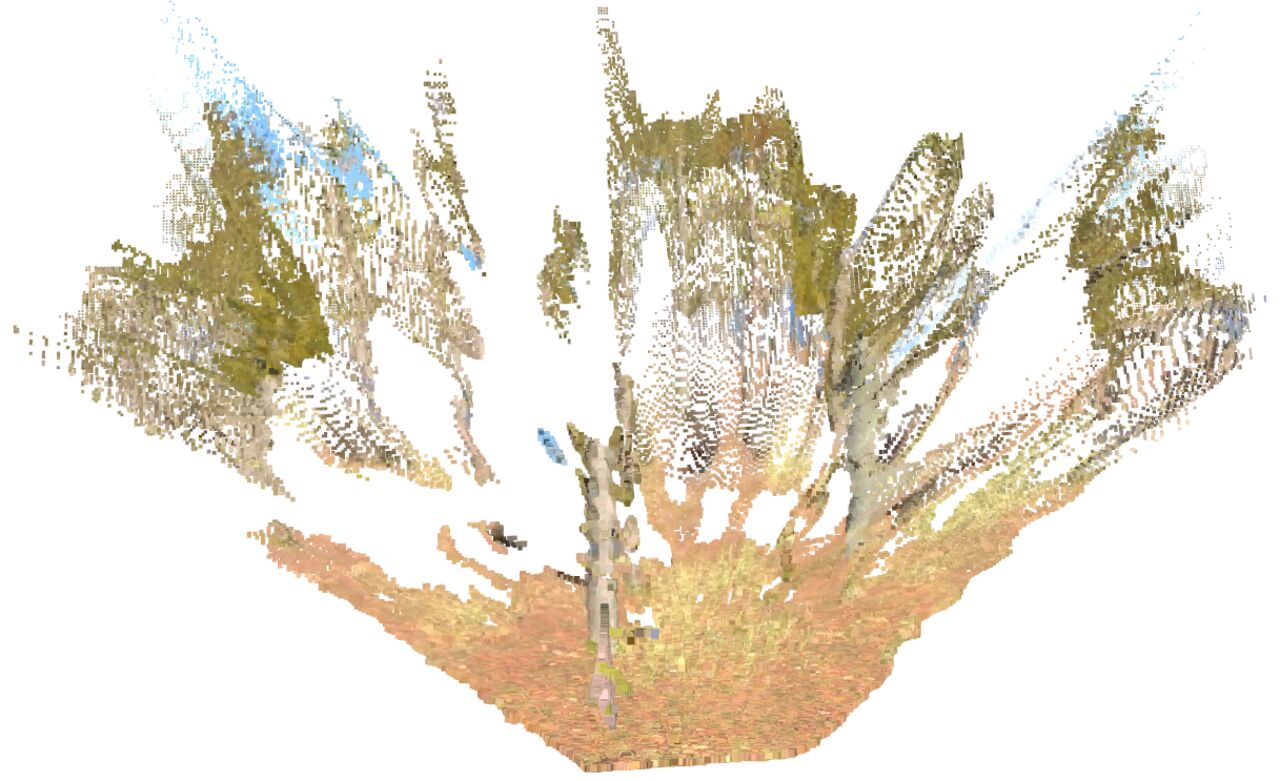}};
        \begin{scope}[x={(a.south east)},y={(a.north west)}]
          \fill[white] (0.001, 0.001) rectangle (0.16,0.21);
          \fill[draw=black, draw opacity=0.5, fill opacity=0] (0,0) rectangle (1, 1);
          \draw (0.08,0.10) node [text=black] {\small (b)};
        \end{scope}
      \end{tikzpicture}
      }%
      \subfloat {\begin{tikzpicture}
        \node[anchor=south west,inner sep=0] (a) at (0,0) {\includegraphics[width=0.155\textwidth]{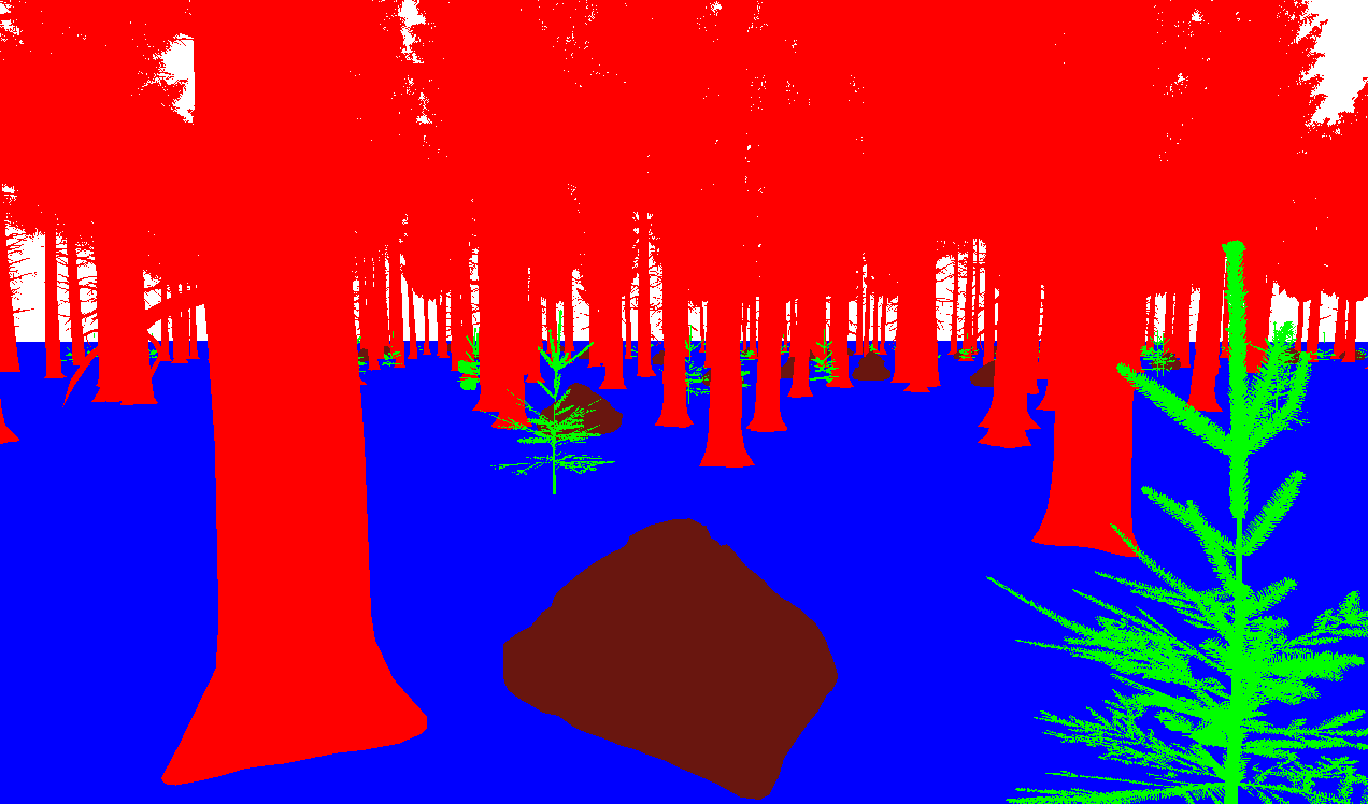}};
        \begin{scope}[x={(a.south east)},y={(a.north west)}]
          \fill[white] (0.001, 0.001) rectangle (0.16,0.21);
          \fill[draw=black, draw opacity=0.5, fill opacity=0] (0,0) rectangle (1, 1);
          \draw (0.08,0.10) node [text=black] {\small (c)};
        \end{scope}
      \end{tikzpicture}}
      \subfloat {\begin{tikzpicture}
        \node[anchor=south west,inner sep=0] (a) at (0,0) {\includegraphics[width=0.155\textwidth]{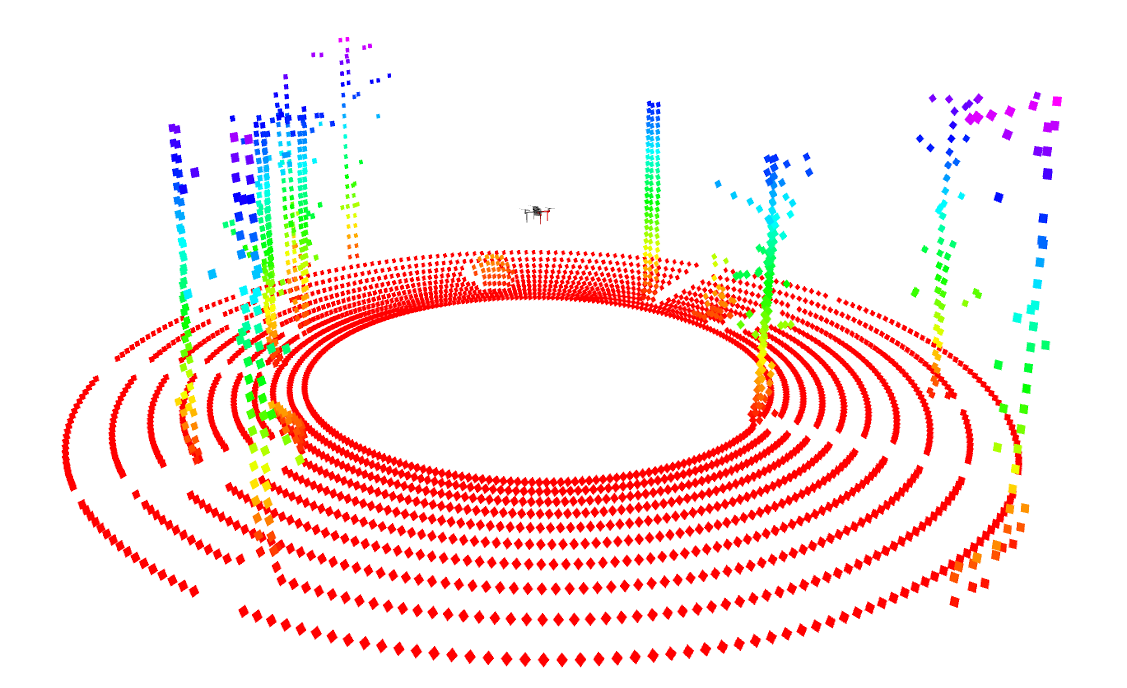} };
        \begin{scope}[x={(a.south east)},y={(a.north west)}]
          \fill[white] (0.001, 0.001) rectangle (0.16,0.21);
          \fill[draw=black, draw opacity=0.5, fill opacity=0] (0,0) rectangle (1, 1);
          \draw (0.08,0.10) node [text=black] {\small (d)};
        \end{scope}
      \end{tikzpicture}
      }%
      \subfloat {\begin{tikzpicture}
        \node[anchor=south west,inner sep=0] (a) at (0,0) {\includegraphics[width=0.155\textwidth]{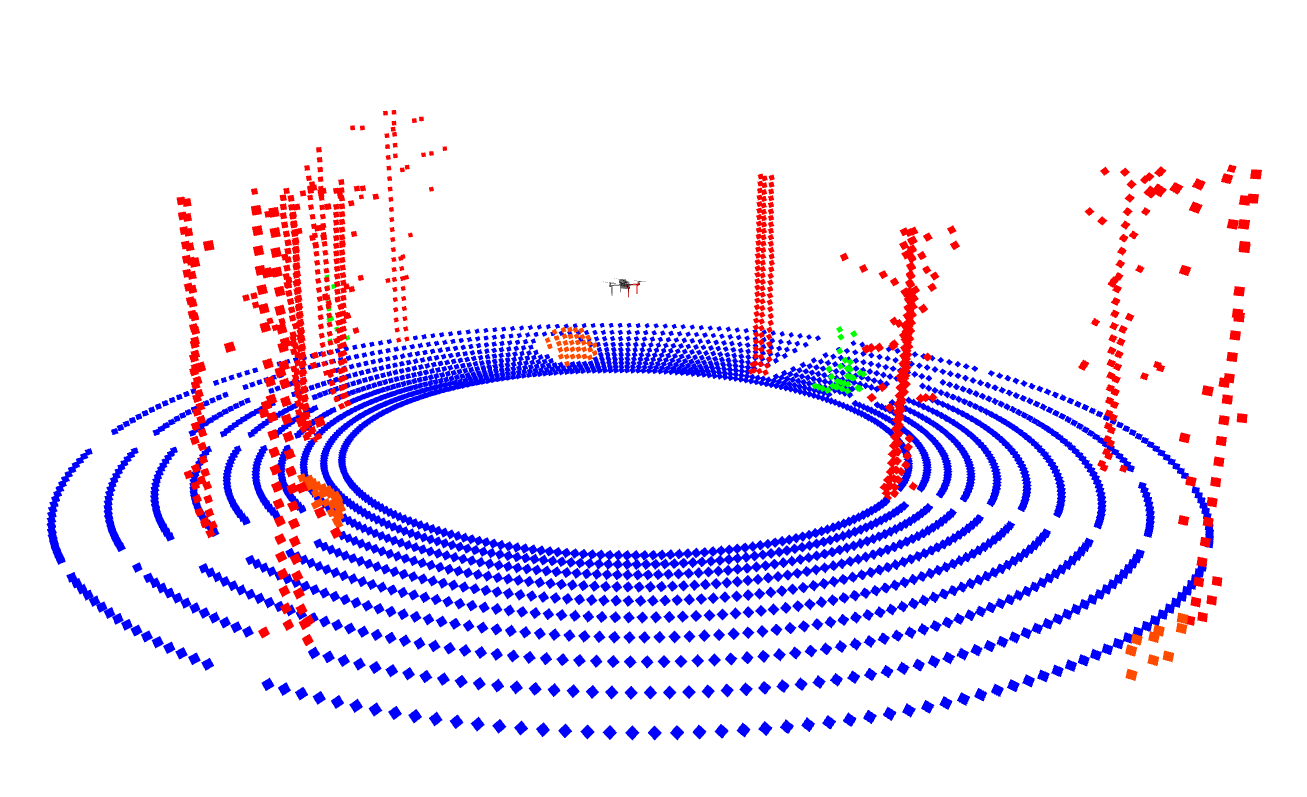}};
        \begin{scope}[x={(a.south east)},y={(a.north west)}]
          \fill[white] (0.001, 0.001) rectangle (0.16,0.21);
          \fill[draw=black, draw opacity=0.5, fill opacity=0] (0,0) rectangle (1, 1);
          \draw (0.08,0.10) node [text=black] {\small (e)};
        \end{scope}
      \end{tikzpicture}
      }%
      \subfloat {\begin{tikzpicture}
        \node[anchor=south west,inner sep=0] (a) at (0,0) {\includegraphics[width=0.155\textwidth]{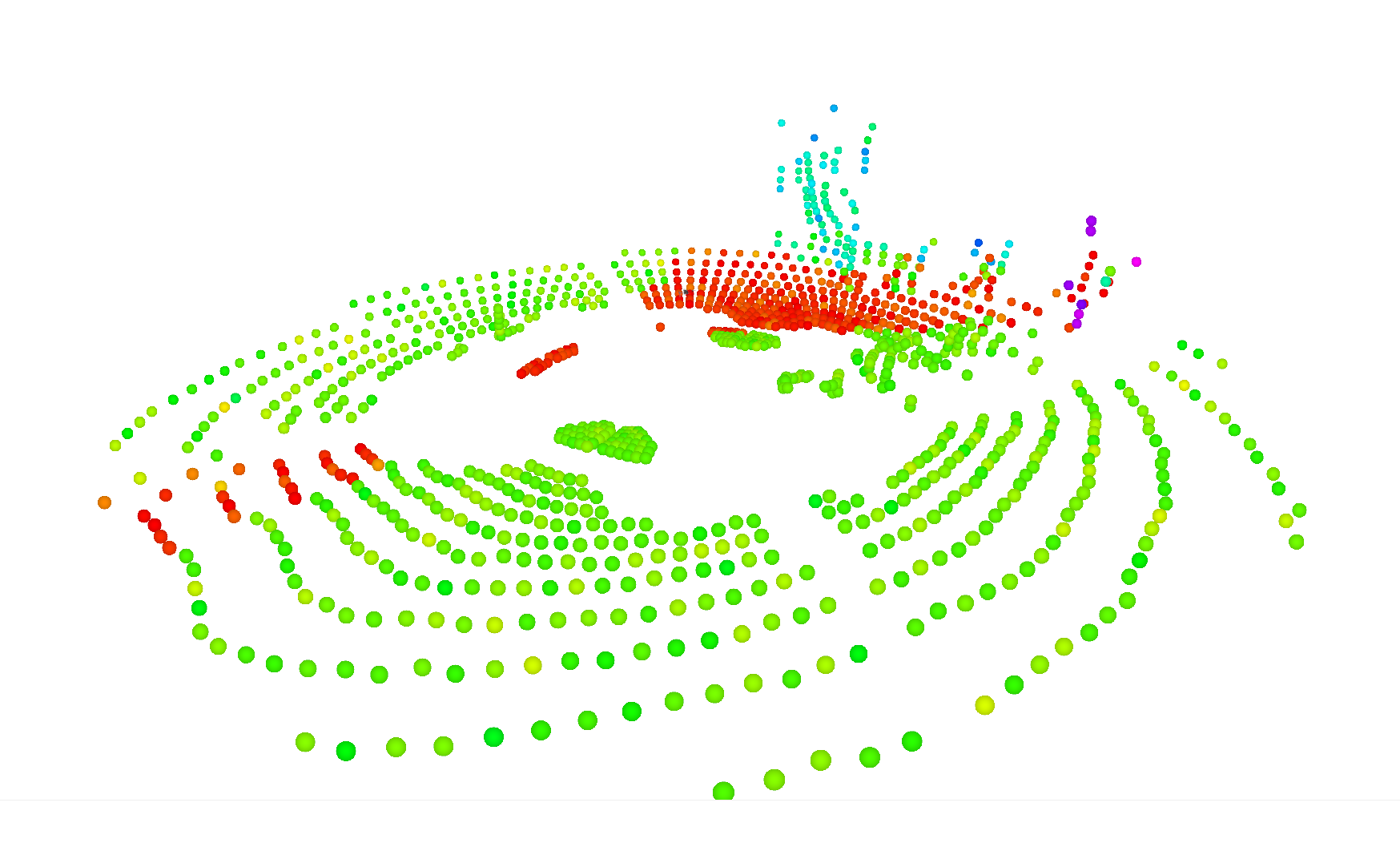}};
        \begin{scope}[x={(a.south east)},y={(a.north west)}]
          \fill[white] (0.001, 0.001) rectangle (0.16,0.21);
          \fill[draw=black, draw opacity=0.5, fill opacity=0] (0,0) rectangle (1, 1);
          \draw (0.08,0.10) node [text=black] {\small (f)};
        \end{scope}
      \end{tikzpicture}
      } \\
      \vspace{-1.1em}
      \caption{The camera sensor outputs: a) RGB, b) stereo depth camera shown as a colored \acp{pc} and c) semantic segmentation mode. The \ac{lidar} sensor outputs: d) a traditional \acp{pc} with a z-coordinate color scale, e) semantic segmentation mode, f) \acp{pc} with intensity information.
      }
      \label{fig:sensors}
      \vspace{-1.8em}
    \end{figure*}

    %%}

    %%{ uav dynamics simulation
    
    \subsection{UAV dynamics simulation}
    
    A single-\ac{uav} simulation can run with real-time factor 1.0 at \SI{30}{\kilo\hertz}, while interacting with the user over \ac{ros}.
    When using the simulation directly by including the header-only implementation, the simulation rate can be even higher.
    Additionally, 400 \ac{uav}s can be simulated in real-time with a dynamics simulation at \SI{250}{Hz}.

    \subsection{Sensor suite}

    The sensor suite of the simulator is capable of producing high-fidelity data that is comparable to the outputs of the real sensors.
    The camera sensor is capable of producing high-resolution images with realistic lighting, shadows, and textures.
    In Figure\, \ref{fig:sensors}, the camera sensor output is shown in RGB, stereo depth, and semantic segmentation mode.
    We test the RGB camera sensor's frame rate in the available environments and with different resolutions.
    A comparison with Flightmare~\cite{song2021flightmare} of the RGB camera rendering speed in frames per second for various camera resolutions is shown in Table\,~\ref{tab:comparison_fps}(a).
    Our environment used for the comparison is the \emph{Warehouse} and \emph{Garage} in the case of Flightmare.

       %%{ TABLE FPS COMPARISON

    \begin{table}[htb!]
      \centering
      \caption{Comparison of the RGB camera rendering speeds (a) in frames per second for various resolutions, and comparison of the frame rate of the \ac{lidar} sensor (b) with varying number of \acp{pc} points.
      \vspace{-1.3em}
      }
      \begin{subtable}{.65\linewidth}
          {\renewcommand{\tabcolsep}{3.5pt}
           \renewcommand{\arraystretch}{0.6}
           \centering
           \caption{\vspace{-0.5em}}
          \begin{tabular}{cccc}
            \toprule
            {\bf Resolution} & {\bf FOV} & {\bf Flightmare~\cite{song2021flightmare}} & {\bf Ours} \\
            \midrule
            32$\times$32 & 120 & \textbf{167} & 117\\
            \midrule
            128$\times$128 & 120 & \textbf{115} & 95\\
            \midrule
            640$\times$480 & 120 & 26 & \textbf{56}\\
            \midrule
            1280$\times$720 & 120 & 23 & \textbf{38}\\
            \bottomrule
          \end{tabular}
          }
      \end{subtable}%
      \begin{subtable}{.33\linewidth}
      
      {\renewcommand{\tabcolsep}{3.5pt}
           \renewcommand{\arraystretch}{0.6}
            \centering
            \caption{\vspace{-0.5em}}
              \begin{tabular}{cc}
                \toprule
                \textbf{No. of points} & \textbf{FPS} \\
                \midrule
                %64 & 9779 \\
                128 & 7646 \\
                256 & 5509 \\
                1024 & 1899 \\
                4096 & 517 \\
                8192 & 257 \\
                32768 & 69 \\
                \bottomrule
            \end{tabular}
            }
     \end{subtable}%
    \label{tab:comparison_fps}
    \end{table}

    %%}
    
    %%{ LIDAR EVALUATION TABLE
    
%  \begin{table}[htb!]
%     \centering
%     \caption{The frame rate of the \ac{lidar} sensor with varying number of points in the point cloud.}
%     \begin{tabular}{cc}
%         \toprule
%         \textbf{Number of points} & \textbf{FPS} \\
%         \midrule
%         16 & 15818 \\
%         64 & 9779 \\
%         128 & 7646 \\
%         256 & 5509 \\
%         512 & 3455 \\
%         1024 & 1899 \\
%         2048 & 1013 \\
%         4096 & 517 \\
%         8192 & 257 \\
%         16384 & 129 \\
%         32768 & 69 \\
%         \bottomrule
%     \end{tabular}
%     \label{tab:lidar_fps}
% \end{table}

    %%}

    The \ac{lidar} sensor is capable of producing a 3D \acp{pc} representation of the environment.
    The measurements do not suffer from artifacts, as in the case of the \ac{lidar} sensor in Gazebo simulator.
    In Figure\, \ref{fig:sensors}, the traditional z-coordinate, the semantic segmentation, and the intensity \acp{pc} are shown.
    \ac{lidar} sensor's frame rate was tested with a varying number of points in the \acp{pc}.
    The results are shown in Table\,~\ref{tab:comparison_fps}(b).

    \subsection{Integration with the MRS UAV System}

\begin{figure}[htb!]
      \centering
      \vspace{-0.0em}
      \subfloat {\begin{tikzpicture}
        \node[anchor=south west,inner sep=0] (a) at (0,0) {\includegraphics[width=0.40\textwidth]{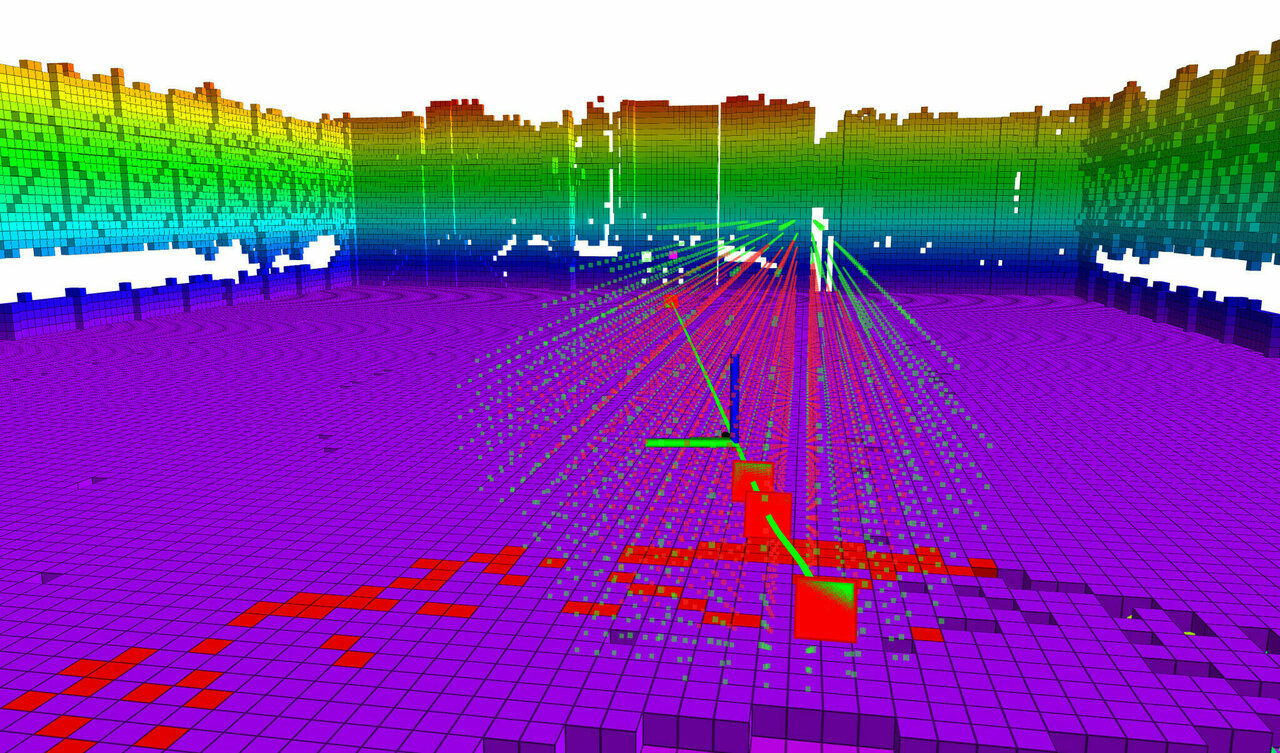}};
        \begin{scope}[x={(a.south east)},y={(a.north west)}]
          %%{ grid
          % % useful grid to help you find coordinates for plotting the overlay
          % \draw[black, xstep=.1, ystep=.1] (0,0) grid (1,1);
          % \foreach \i in {0,0.1,0.2,0.3,0.4,0.5,0.6,0.7,0.8,0.9,1} {
          %   \node[align=center] at (\i, -0.05) {\i};
          %   \node[align=center] at (\i, 1.05) {\i};
          %   \node[align=center] at (-0.05, \i) {\i};
          %   \node[align=center] at (1.05, \i) {\i};
          % }
          %%}
          % plot some stuff over the image
          % \fill[white] (0.001, 0.001) rectangle (0.05,0.08);
          \fill[draw=black, draw opacity=0.5, fill opacity=0] (0,0) rectangle (1, 1);
          % \draw (0.025,0.035) node [text=black] {\small (c)};
        \end{scope}
      \end{tikzpicture}} \\
      \vspace{-0.5em}
      % \caption{Showcase of the MRS UAV System's \acl{hla}: a) RGB image from the onboard camera, b) map of the \emph{warehouse} environment within the \emph{OctoMap framework} after exploration, c) visualization of the planner's expansion through the environment.}
      \caption{Showcase of the MRS UAV System's \acl{hla}: map of the \emph{warehouse} environment within the \emph{OctoMap framework} after exploration and the visualization of the planner's expansion through the environment.}
      \label{fig:exploration}
      \vspace{-1em}
    \end{figure}

    The \ac{ue5} simulator offers seamless (but optional) integration with the MRS UAV System \cite{baca2021mrs}.
    The system provides users a variety of high-level and low-level features.
    Users can implement their own plugins for state estimators, feedback and feedforward controllers, reference generators, trajectory planners, and mappers, without the need to modify the underlying system.
    Therefore, the users can benefit from the rest of the ecosystem while focusing on their particular field of research.
    Moreover, the MRS system provides a complete 3D exploratory navigation pipeline, which includes visual and \ac{lidar} \acp{slam}, occupancy mapper and planner.
    Figure\,\ref{fig:exploration} shows the results of full autonomy exploration through the \emph{warehouse}.
    These features are available to the users out of the box.

\begin{figure}[htb!]
      \centering
      \includegraphics[width=0.45\textwidth]{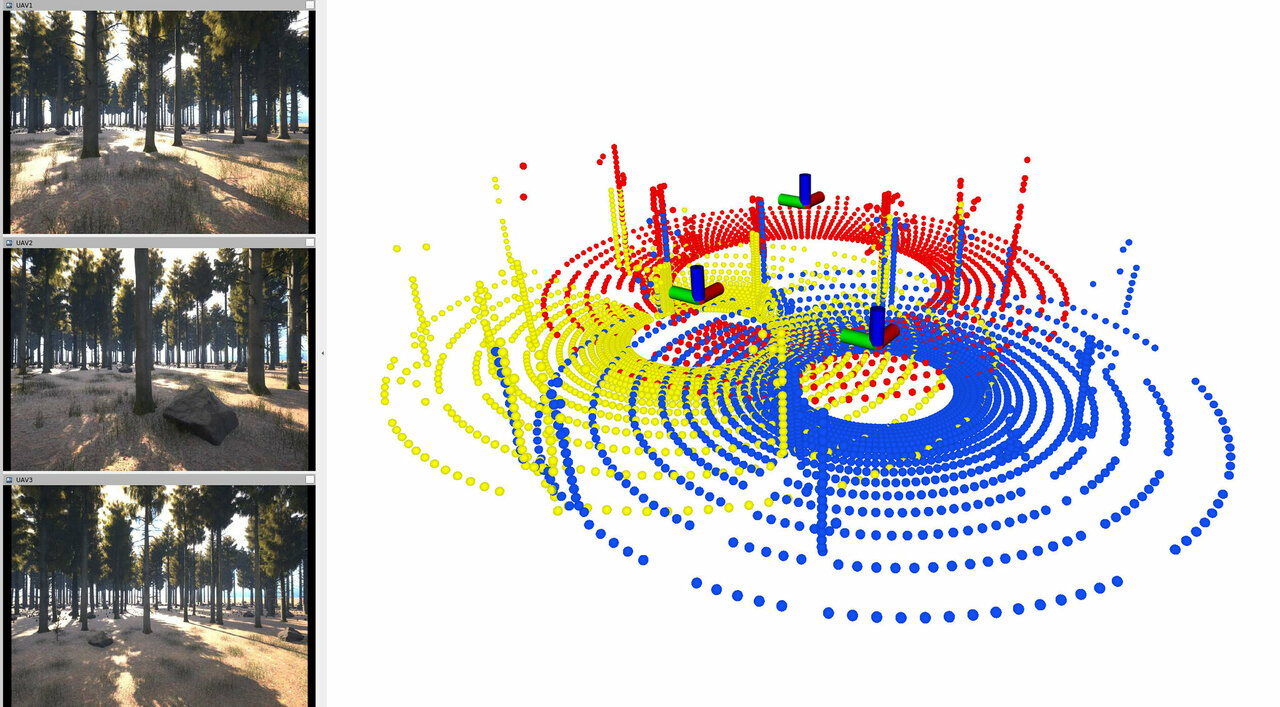}
      \vspace{-0.7em}
      \caption{Showcase of a multi-UAV simulation: 3 \acp{uav} equipped with RGB cameras and \acp{lidar}}.
      \label{fig:multi_uav_sensors}
      \vspace{-1.7em}
    \end{figure}

    Moreover, as the name suggests, the MRS UAV System and the proposed \ac{ue5} simulator support multi-robot simulations.
    Simple parametrization in the \ac{ros} interface to the simulator allows adding additional drones to the simulation and specifying their initial conditions and sensor configuration.
    \emph{The dynamic reconfigure server} within \ac{ros} provides options for adjusting the sensor settings on the fly.
    Figure\,\ref{fig:multi_uav_sensors} shows visualization from such a multi-UAV simulation in \emph{Forest} environment created during flight by the proposed novel procedural generation.

    %%{ FIGURE NAVIGATION

    %%}

    %%{ Figure multi-uav

    %%}

    %%}

  %%{ Conclusion
%that integrates high-level autonomy (e.g., planning, control and mapping), and it enables simulations of complex tasks, e.g. autonomous navigation in unknown cluttered environments.
  
  \section{Conclusion}

  FlightForge is the novel \acl{ue5}-based simulator for \acp{uav}.
  The key contributions of our work include advanced rendering capabilities, simulation of sensors, diverse control modalities for UAVs, and integration with the fully autonomous MRS UAV System.
  Unlike existing simulators, FlightForge offers a solution that encompasses high-fidelity simulation with procedural environment generation and a fully autonomous UAV system.
  The simulator supports single and multi-UAV deployment, dynamic real-time sensors equipped with annotated semantic segmentation ground truth.
  Through experimental validation, we have demonstrated the efficacy of our framework in facilitating autonomous UAV navigation across a large range of environments.
  Its ease of installation via Docker, and Apptainer container systems ensures accessibility for researchers and developers.
  Future work includes expanding the sensor suite, and completing the migration from ROS\,1 to ROS\,2.
  Our proposed framework serves as a valuable tool for accelerating progress in autonomous systems and paves the way for future advancements in robotics.

  %%}

  \bibliographystyle{IEEEtran}
  \bibliography{ref.bib}

\end{document}